\documentclass[10pt,onecolumn,letterpaper]{article}

\usepackage{times}
\usepackage{epsfig}
\usepackage{graphicx}
\usepackage{amsmath}
\usepackage{amssymb}
\usepackage{epsfig}
\usepackage{times}
\usepackage{graphicx}
\usepackage{subfigure}
\usepackage{color}
\usepackage{amsthm}
\usepackage{algorithm}
\usepackage{algorithmic}
\usepackage{wrapfig}
\usepackage{url}
\usepackage{multirow}
\makeatletter
\renewcommand{\@thesubfigure}{}
\makeatother

\newcommand{\Figure}[1]{Fig.~{\color{red}#1}}
\newcommand{\Figures}[2]{Figs.~#1 and~#2}

\newcommand{\Equation}[1]{Eqn.~(#1)}

\newcommand{\MyCaption}[1]{\caption{{#1}}}
\newcommand{\Mytextquotedbl}[1]{{\textquotedblleft}#1\textquotedblright}
\newcommand{\mytextbf}[1]{\vspace{.1cm}\noindent\textbf{{#1}}\vspace{.1cm}\\}

\newcommand{\Table}[1]{Tab.~#1}

\providecommand{\myproofname}{\footnotesize{Proof}}

\begin{document}

%%%%%%%%% TITLE
\title{Hybrid Affinity Propagation\thanks{A short version appeared in~\cite{XuWHL11}}}

\author{Jingdong Wang\footnotemark[2]~~~~Hao Xu\footnotemark[3]~~~~Xian-Sheng Hua\footnotemark[2]~~~~Shipeng Li\footnotemark[2]\\
\footnotemark[2]~Microsoft Research Asia~~~~~~\footnotemark[3]~University of Science and Technology of China~~~~~~\\
{\tt\small \{jingdw, xshua, spli\}@microsoft.com}~~~
{\tt \small xuhao657@ustc.edu}
}
\date{~}
\maketitle

%%%%%%%%% ABSTRACT
\begin{abstract}
In this paper,
we address a problem of managing tagged images
with hybrid summarization.
We formulate this problem
as finding a few image exemplars
to represent the image set
semantically and visually,
and solve it
in a hybrid way
by exploiting
both visual and textual information associated with images.
We propose a novel approach,
called homogeneous and heterogeneous message propagation ($\text{H}^\text{2}\text{MP}$).
Similar to the affinity propagation (AP) approach,
$\text{H}^\text{2}\text{MP}$
reduce the conventional~\emph{vector} message propagation
to~\emph{scalar} message propagation
to make the algorithm more efficient.
Beyond AP
that can only handle homogeneous data,
$\text{H}^\text{2}\text{MP}$ generalizes it
to exploit extra heterogeneous relations
and the generalization is non-trivial
as the reduction to scalar messages
from vector messages
is more challenging.
The main advantages of our approach
lie in
1) that $\text{H}^\text{2}\text{MP}$ exploits
visual similarity
and in addition the useful information from the
associated tags, including the associations relation between
images and tags and the relations within tags,
and 2) that the summary is both visually and semantically satisfactory.
In addition,
our approach can also present a textual summary
to a tagged image collection,
which can be used to automatically
generate a textual description.
The experimental results
demonstrate the effectiveness and efficiency
of the proposed approach.
\end{abstract}

\section{Introduction}
\label{sec:introduction}
The increasing development of image search engines,
photo-sharing web sites,
and desktop photo management tools,
has made people
easily access a large amount of images.
However, image collections
are usually unorganized,
which
makes finding desired photos
and quick overview of an image collection
very difficult.
This unstructured nature of image collections
has attracted
great effort
on computing visual summaries.
On the other hand,
most image collections
are provided
with rich text information,
and such image collections are called tagged image collections in this paper.
For example,
images on Flickr
are titled, tagged, and commented by users.
Images from the Web
are often associated with surrounding texts.
The text information usually reflects the semantic content of images
and is helpful for summarization.

\begin{figure*}
\centering
(a)~~{\includegraphics[height = .15\linewidth]{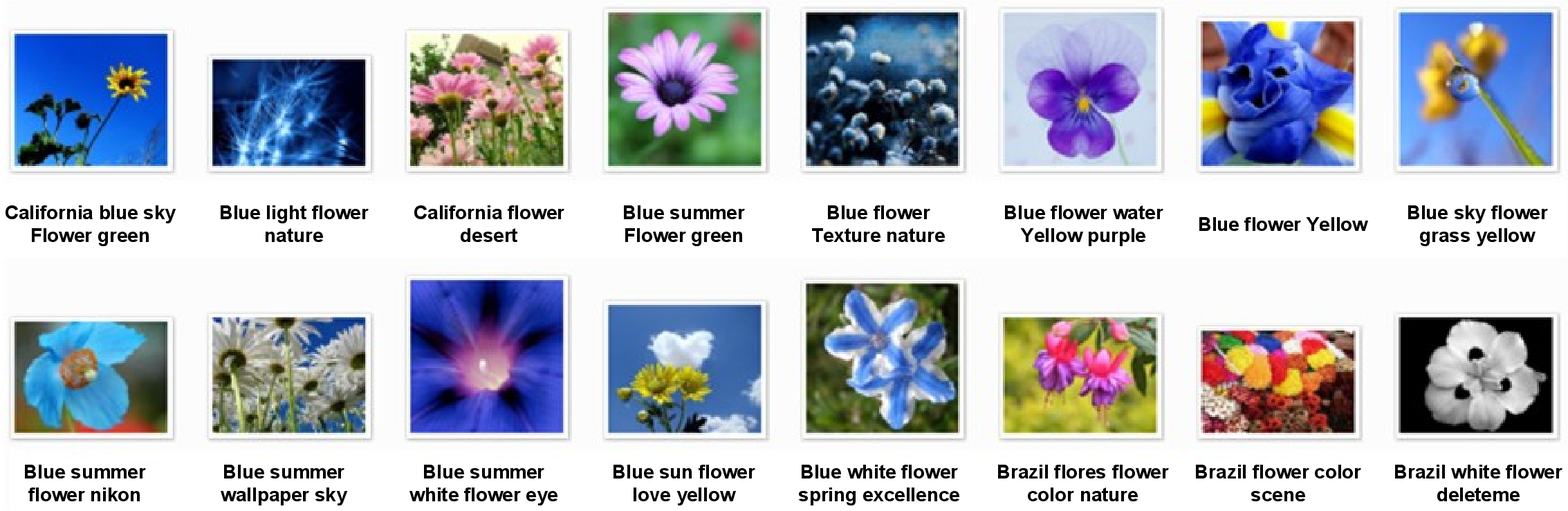}}~~~~~~~~
(b)~~{\includegraphics[height = .15\linewidth]{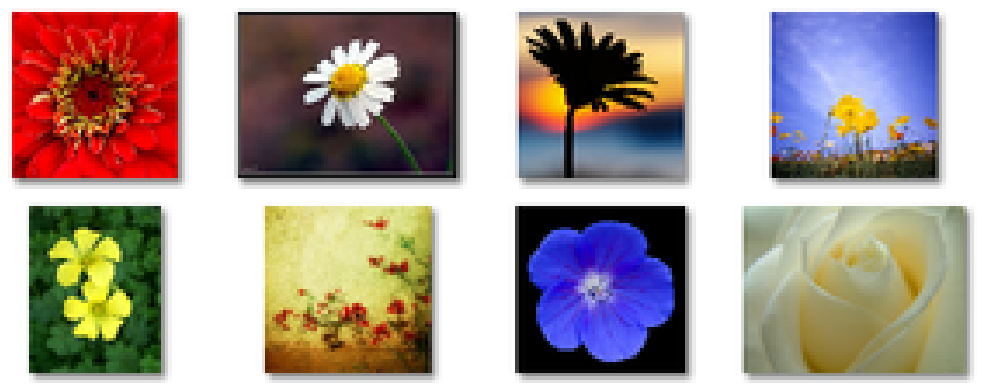}}
\MyCaption{An example of a visual summary for an image collection.
(a) shows randomly selected images and their associated texts
from the input, a set of tagged images,
and (b) shows its summary
identified by our hybrid summarization scheme.
}
\label{fig:teaser}
\end{figure*}

In this paper,
we address the image management task
through a hybrid summarization scheme.
The key is to find the summary
in a hybrid way
to exploit both visual and textual information.
An example is shown in~\Figure{\ref{fig:teaser}}.
Given rich tag information associated with images,
there are three useful relations from images and tags:
two homogeneous relations within images and tags,
including image similarity and tag similarity,
and one heterogeneous relation between images and tags,
e.g., their association relations.
We propose a hybrid summarization approach
to find image exemplars
through investigating all three relations together
including the information from the
associated tags, i.e.,
association relations between
images and tags and relations within tags
so that the summary is both visually and semantically satisfactory.

\subsection{Related work}
Most existing image sharing web sites present an overview
of an image collection
by showing the top images (e.g., Flickr group~\cite{FlickrGroup}),
which obviously does not present a good summary,
or allowing consumers to manually select images (e.g., Picassa web album~\cite{PicasaWebAlbum}),
which is inconvenient for consumers particularly in a large number of images.

Rother et al.~\cite{RotherKKB05} summarize a set of images with
a~\Mytextquotedbl{digital tapestry}.
They synthesize a large output image
from a set of input images, stitching together salient and
spatially compatible blocks from the input images.
Wang et al.~\cite{WangQSTS06} create a~\Mytextquotedbl{picture collage},
a 2D spatial arrangement
of the images in the input set chosen to maximize visible salient regions.
These works do not address the problem of selecting the set of images
to appear in the summary.

Recently, there are a few works
to deal with the selection problem.
Simon et al.~\cite{SimonSS07}
selects a set of images
using the greedy k-means algorithm,
by examining the distribution of images
to select a set of canonical views
only based on visual features
without exploiting the associated tags.
Raguram and Lazebnik~\cite{RaguramL08}
select iconic images to summarize general visual categories
using a simple joint clustering technique
from both appearance and semantic aspects.
It first obtained two independent clusters
from the visual feature and the textual feature, respectively,
and then takes their intersection to get the final clustering,
but the joint process is obtained sequentially
instead of simultaneously.
Surrounding texts are limitedly exploited
for image grouping~\cite{GaoLQZCM05, RegeDH08}
by considering the association relations between words and images
using the co-clustering technique,
but without investigating interior relations over tags.

Image summarization is also studied
in the information retrieval community.
Clough et al.~\cite{CloughJS05} construct a hierarchy of images
using only textual caption data, and the concept of subsumption.
Schmitz~\cite{Schmitz06} uses a similar approach
but relies on Flickr tags.
Jaffe et al.~\cite{JaffeNTD06}
summarize a set of images
using only tags and geotags.
By detecting correlations
between tags and geotags,
they are able to produce tag maps,
where tags and related images are overlaid
on a geographic map
at a scale corresponding to the range
over which the tag commonly appears.
All these approaches
could be used to further organize the images.
However, none of them
exploits the visual information.

\subsection{Our approach}
In this paper,
we present a hybrid summarization approach
to find a few image exemplars
to represent the image collection,
which is both visually and semantically satisfactory.
Toward this end,
we propose an effective scalar hybrid message propagation scheme
over images and tags,
homogeneous and heterogeneous message propagation ($\text{H}^\text{2}\text{MP}$),
to exploit simultaneously
homogeneous relations within images and tags
and heterogeneous relations between images and tags.
It is beyond the affinity propagation algorithm~\cite{FreyD07}
that only can handle homogeneous data,
and $\text{H}^\text{2}\text{MP}$ can effectively exploit the heterogeneous relations
between images and tags
as well as the interior relations within tags.
Moreover,
the reduction from vector messages to scalar messages
is more challenging
than AP
because $\text{H}^\text{2}\text{MP}$ contains additional heterogeneous relations.
Besides,
our approach is superior over
existing co-clustering algorithms~\cite{Dhillon01,DhillonMM03}
that only utilize the heterogeneous relations
because 1) it directly obtains
the exemplars instead of performing the necessary postprocess
to find the centers followed by a clustering procedure
and 2) it takes advantage of
homogeneous relations with images and tags
as well as heterogeneous relations between them.

\subsection{Notation}
\begin{figure}[t]
\centering
\includegraphics[scale = .6]{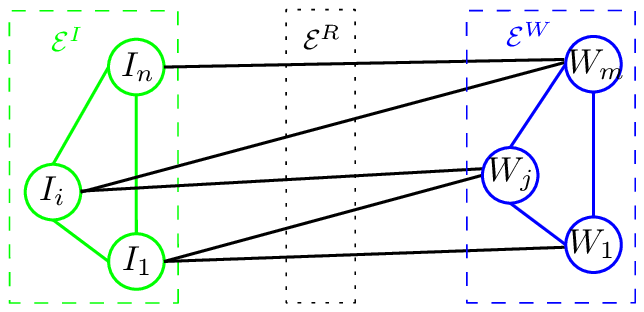}
\MyCaption{Heterogeneous graph over images and tags.}
\label{fig:HeterogeneousGraph}
\end{figure}

Given a set of $n$ images,
$\mathcal{I} = \{I_1, I_2, \cdots, I_n\}$,
a set of corresponding texts,
$\mathcal{T} = \{\mathcal{T}_1, \mathcal{T}_2, \cdots, \mathcal{T}_n\}$,
$\mathcal{T}_k = \{W^k_1, W^k_2, \cdots, W^k_{m_k}\}$,
and
the union set of tags $\mathcal{W} = \{W_1, \cdots, W_m\}
= \mathcal{T}_1\cup\mathcal{T}_2\cup\cdots\cup\mathcal{T}_n$,
we aim to
find a summary,
a set of image exemplars,
$\bar{\mathcal{I}} \in \mathcal{I}$.
There are three types of relations
over images and tags
as depicted in~\Figure{\ref{fig:HeterogeneousGraph}}.
The heterogeneous relations
between all pairs of associated images and tags
are represented
by the edges,
$\mathcal{E}^R$.
The homogeneous relations
within images are
represented
by the edges,
$\mathcal{E}^I$,
and the similarity between a pair of images $i$ and $k$
is denoted by $s'_{I}(i, k)$.
The homogeneous relations
within tags are
represented by the edges,
$\mathcal{E}^W$,
and the similarity between a pair of tags $j$ and $k$
is represented by $s'_{W}(j, k)$.

Suppose a set of image exemplars to be identified
be denoted as
$\bar{\mathcal{I}} = \{ I_{c_1}, I_{c_2}, \cdots, I_{c_n}\}$,
where $c_k \in \{1, 2, \cdots, n\}$ is
the exemplar image index of image $I_k$,
and
$\mathbf{c} = [c_1~c_2~\cdots~c_n]^T$
is a label vector over images.
If such a label vector
satisfies a valid constraint
that an image should also serve as the exemplar of itself if it is an exemplar of any other image,
it would uniquely correspond to a set of image exemplars.
In other words,
identifying the exemplars
can be viewed as searching over valid labels.

\section{Affinity propagation}
\label{sec:Background}
To make our approach easily understood,
we first review affinity propagation (AP)~\cite{FreyD07}.
AP is an approach to find a good subset of exemplars
for a whole set of homogeneous data points,
by considering all data points
as candidate exemplars
such that they can represent the image collection very well,
and mathematically it is formulated to maximize
an objective function,

{\footnotesize
\vspace{-.2cm}
\begin{align}
\mathcal{S}(\mathbf{c})
= E^I(\mathbf{c}) + V^I(\mathbf{c}),
\label{eqn:APobjectivefunction}
\end{align}
\vspace{-.2cm}}

\noindent where $E^I(\mathbf{c})$ is a fitting function
to evaluate how well the image exemplars represent the other images,
and written as

{\footnotesize
\vspace{-.2cm}
\begin{align}
E^I(\mathbf{c}) = \sum\nolimits_{i = 1}^{n} s'_{I}(i, c_i),
\end{align}
\vspace{-.2cm}}

\noindent and $V^I(\mathbf{c})$ is a valid configuration function
to constrain that
an image must select itself as its exemplar
if it is selected as an exemplar of other data point,
and it is formulated as

{\footnotesize
\vspace{-.2cm}
\begin{equation}
V^I(\mathbf{c})
= \sum\nolimits_{k = 1}^{n} \delta_k(\mathbf{c}),
\end{equation}}
\noindent where
{\footnotesize
\begin{equation}
\delta_k(\mathbf{c}) =
\left\{
\begin{array}{ll}
-\infty, & \text{if~} c_k \neq k \text{~but~}\exists i: c_i = k\\
0, & \text{otherwise}.
\end{array}\right.
\label{eqn:tp:priormodel}
\nonumber
\end{equation}
\vspace{-.2cm}}

\noindent This objective function can be depicted
by a factor graph
over the variables $\{c_i\}$,
and the function terms $\{\delta_i\}$ and $\{s_{i}\}$,
which correspond to the subgraph in the dashed green box
in~\Figure{\ref{fig:FactorGraphForOverallObjective}}.
AP to maximize~\Equation{\ref{eqn:APobjectivefunction}} is a scalar message propagation algorithm
and derived from the max-sum algorithm,
which transits two vector-messages
between $c_i$ and $\delta_k$.
The message, $\rho_{i \rightarrow k}$,
sent from $c_i$ to $\delta_k$,
consists of $n$ real numbers,
with one for each possible value of $c_i$.
The message,
$\alpha_{i\leftarrow k}$,
sent from $\delta_k$ to $c_i$,
also consists of $n$ real numbers.
For simplicity,
the subscript $I$ may be dropped in the following presentation.
The two messages are depicted
in~\Figures{\ref{fig:Messages:rho}}{\ref{fig:Messages:alpha}}
neglecting the massages corresponding to the red edges,
and formulated as follows,

{\footnotesize
\vspace{-.2cm}
\begin{equation}
\rho_{i \rightarrow k}(c_i)
= s(i, c_i) + \sum\nolimits_{k': k'\neq k } \alpha_{i\leftarrow k'}(c_i),
\end{equation}
\begin{align}
\alpha_{i \leftarrow k} (c_i)
=& \max_{h_1, \cdots, h_{i-1}, h_{i+1}, \cdots, h_n}[
\sum\nolimits_{i': i' \neq i} \rho_{i' \rightarrow k} (h_{i'}) \nonumber \\
&+ \delta_k(h_1, \cdots, h_{i-1}, c_i, h_{i+1}, \cdots, h_n)].
\end{align}
\vspace{-.2cm}
}

As shown in~\cite{FreyD07},
this vector-message propagation scheme
can be reduced to a scalar-message propagation scheme
between data points.
There are two kinds of messages
exchanged within image points.
The~\Mytextquotedbl{responsibility}
$r(i,k)$,
sent from data point $i$ to data point $k$,
which reflects how well $k$ serves as the exemplar of $i$
considering other potential exemplars for $i$,
and the~\Mytextquotedbl{availability}
$a(i,k)$,
sent from data point $k$ to data point $i$,
which reflects how appropriately $i$ chooses $k$
as its exemplar considering other potential points that may choose $k$ as their exemplar.
The messages are updated in an iterative way as

{\footnotesize
\vspace{-.2cm}
\begin{equation}
r(i, k) = s(i, k) - \max\nolimits_{i': i' \neq k}[s(i, i') + a(i, i')].
\label{eqn:RMessage}
\end{equation}
\begin{equation}
a(i, k) = \left\{
\begin{array}{ll}
\sum\nolimits_{i': i' \neq k} \max(0, r(i', k)) & k = i, \\
\min[0, r(k, k) + \sum\nolimits_{i': i' \neq i, k} \max(0, r(i', k))] & k \neq i.
\end{array}\right.
\label{eqn:AMessage}
\end{equation}
\vspace{-.2cm}
}

\section{Hybrid summarization}
\label{sec:MainSection}
Affinity propagation has been shown to be very effective to find exemplars~\cite{FreyD07},
but it can only handle homogeneous data points.
In the following,
we present a new hybrid message propagation approach,
to generalize AP to heterogeneous data points,
homogeneous and heterogeneous message propagation ($\text{H}^\text{2}\text{MP}$).
$\text{H}^\text{2}\text{MP}$ is applied
to find image exemplars
for a tagged image collection,
with the advantages of exploiting not only the visual information of images
but also heterogeneous relations between images and their associated tags
and interior similarities within words
in a hybrid way.

We exploit the associated tag information for image identification
by augmenting image exemplar identification
with tag exemplar identification
and bridging image exemplars and tag exemplars
according to association relations
between images and tags.
Thus,
we define a label vector over tags
$\mathbf{b} = [b_1~b_2~\cdots~b_m]^T$
to represent the tag exemplars,
i.e.,
$W_j$ selects $W_{b_j}$ as its exemplar.
Basically,
the proposed approach
is modeled according to the following two properties:
1) these image and tag exemplars
are good representatives of images and tags,
respectively,
and 2)
these image and tag exemplars
reflect their association relations.
The second property investigates the association relations between images and tags
and also serves as a bridge
to make use of the relations within tags.

The first property concerns
how well image and word exemplars
represent the other images and tags
if only the homogeneous relations are
taken into consideration.
For images,
this is formulated as $E^I(\mathbf{c})$
in~\Equation{\ref{eqn:APobjectivefunction}},
and for tags
we can get a similar formulation,

{\footnotesize
\vspace{-.2cm}
\begin{align}
E^W(\mathbf{b}) = \sum\nolimits_{j = 1}^{m} s'_{W}(j, b_j).
\end{align}
\vspace{-.2cm}
}

%\begin{figure}[t]
%\centering
%\includegraphics[scale = .6]{fig/bifactorgraph}
%\MyCaption{Factor graph for
%the overall objective function~\Equation{\ref{eqn:OverallObjectiveFunction}}.
%$\square$ represents a function node,
%and $\bigcirc$ represents a variable node.}
%\label{fig:FactorGraphForOverallObjective}
%\end{figure}

\begin{figure}[t]
\centering
\includegraphics[scale = 1.]{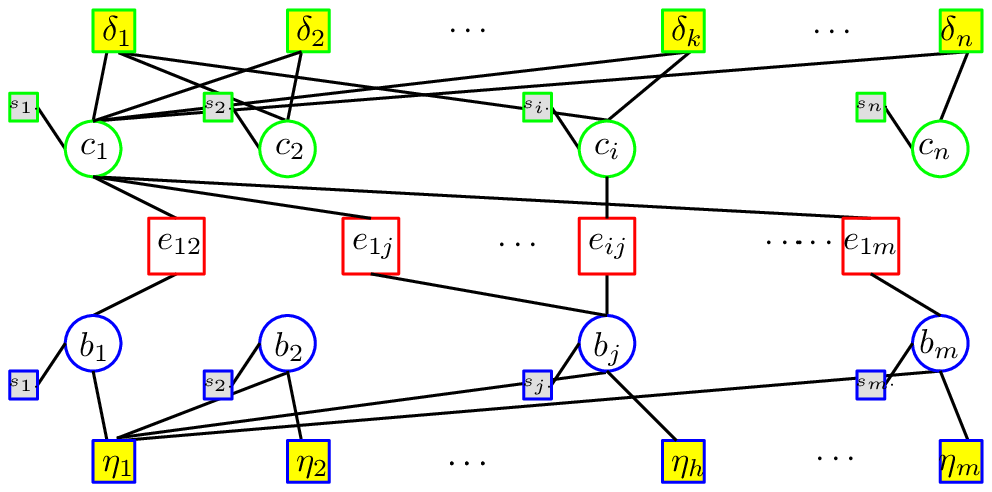}
\MyCaption{Factor graph for
the overall objective function~\Equation{\ref{eqn:OverallObjectiveFunction}}.
$\square$ represents a function node,
and $\bigcirc$ represents a variable node.}
\label{fig:FactorGraphForOverallObjective}
\end{figure}

The second property essentially investigates
the effect of
the heterogeneous relations
between images and tags
on exemplar identification,
and serves as a bridge
to get help for image exemplar identification
from tag information.
We would like to
assign different preferences
for a pair of connected image $i$ and tag $j$
according to
whether image $i$ or tag $j$ selects itself as its exemplar.
This affect is formulated
as a function $e_{ij}(c_i, b_j)$ over a pair of image and tag
$(i, j)$
and their exemplars
$(c_i, b_j)$,
$e_{ij}(c_i, b_j)$.
The whole affect function is written
as follows,

{\footnotesize
\vspace{-.2cm}
\begin{equation}
R(\mathbf{c}, \mathbf{b})
= \sum\nolimits_{(i, j) \in \mathcal{E}^R} e_{ij}(c_i, b_j),
\vspace{-.2cm}
\end{equation}}

\noindent where $e_{ij}(c_i, b_j)$
aims to set different weights
according to whether $c_i$ is equal to $i$
and whether $b_j$ is equal to $j$,

{\footnotesize
\vspace{-.2cm}
\begin{equation}
e_{ij}(c_i, b_j) =
\left\{
\begin{array}{ll}
q(i, j), & c_i \neq i, b_j \neq j\\
\bar{q}(i, j), & c_i = i, b_j = j\\
p(i, j), & c_i = i, b_j \neq j\\
p(j, i), & c_i \neq i, b_j = j.\\
\end{array}\right.
\label{eqn:tp:priormodel}
\nonumber
\end{equation}
\vspace{-.2cm}}

\noindent In this paper,
we instantiate this affect function
similar to the Ising model
based on the following aspects:
if an image is selected as an exemplar,
the tags linking this image
should have larger probability
to be selected as exemplars,
and vice versa.
In other words,
we would assign a higher penalty
for a pair of related image $i$ and tag $j$
when they do not select themselves as their exemplars simultaneously
or do not select others as their exemplars simultaneously.
Specifically,
we set $p(i,j)$ and $p(j,i)$ to negative values
and set $q(i, j)$ and $\bar{q}(i, j)$  to zero.

For $\mathbf{b}$,
a similar valid constraint is defined as

{\footnotesize
\vspace{-.2cm}
\begin{align}
V^W(\mathbf{c}) = \sum\nolimits_{k = 1}^{m} \eta_k(\mathbf{b}),
\end{align}
\vspace{-.2cm}}

\noindent and $\eta_k(\cdot)$ is defined similarly to $\delta_k(\cdot)$.

In summary,
the overall objective function
is as follows

{\footnotesize
\vspace{-.2cm}
\begin{align}
\mathcal{S}(\mathbf{c}, \mathbf{b})
=& \gamma_I (E^I(\mathbf{c}) + V^I(\mathbf{c}))
+ \gamma_W (E^W(\mathbf{b}) + V^W(\mathbf{b}))
+ R(\mathbf{c}, \mathbf{b}) \nonumber \\
=& \sum\nolimits_{i = 1}^{n} \gamma_I s'_I(i, c_i)
+ \sum\nolimits_{k = 1}^{n} \delta_k(\mathbf{c})
+ \sum\nolimits_{j = 1}^{m} \gamma_W s'_W(j, b_j) \nonumber \\
&+ \sum\nolimits_{k = 1}^{m} \eta_k(\mathbf{b})
+ \sum\nolimits_{(i, j) \in \mathcal{E}^R }e_{ij}(c_i, b_j) \nonumber \\
=& \sum\nolimits_{i = 1}^{n} s_I(i, c_i)
+ \sum\nolimits_{k = 1}^{n} \delta_k(\mathbf{c})
+ \sum\nolimits_{j = 1}^{m} s_W(j, b_j) \nonumber \\
&+ \sum\nolimits_{k = 1}^{m} \eta_k(\mathbf{b})
+ \sum\nolimits_{(i, j) \in \mathcal{E}^R }e_{ij}(c_i, b_j),
\label{eqn:OverallObjectiveFunction}
\end{align}
\vspace{-.2cm}
}

\noindent where $\gamma_I$ and $\gamma_W$ are balance weights,
$s_I(i, c_i) = \gamma_I s'_I(i, c_i)$
and $s_W(j, b_j) = \gamma_W s'_W(j, b_j)$.
Maximizing~\Equation{\ref{eqn:OverallObjectiveFunction}}
may get a byproduct, tag exemplars,
and our approach mainly use them as a bridge
to exploit tag information
to help find image exemplars.
We depict~\Equation{\ref{eqn:OverallObjectiveFunction}}
using a factor graph in~\Figure{\ref{fig:FactorGraphForOverallObjective}}.
Each term in~\Equation{\ref{eqn:OverallObjectiveFunction}}
is represented by a~\emph{function node}
and each label $c_i$ (or $b_i$)
is represented by a~\emph{variable node}.
Edges exist only between function and variable nodes,
and a function node is connected to a variable node
iff its corresponding term depends on the variable.
Heterogeneous relations $e_{ij}$
serve as a bridge
to connect two factor graphs over images and tags.

\subsection{$\text{H}^\text{2}$ message propagation}
%\begin{figure} [t]
%\centering
%\subfigure[(a) $\rho_{i\rightarrow k}$]{
%\label{fig:Messages:rho}
%\includegraphics[scale = .7]{fig/rho}
%}
%\subfigure[(b) $\alpha_{i\leftarrow k}$]{
%\label{fig:Messages:alpha}
%\includegraphics[scale = .7]{fig/alpha}
%}
%\subfigure[(c) $\pi_{i\rightarrow e}$]{
%\label{fig:Messages:pi}
%\includegraphics[scale = .7]{fig/pi}
%}
%\subfigure[(d) $\upsilon_{i\leftarrow e}$]{
%\label{fig:Messages:upsilon}
%\includegraphics[scale = .7]{fig/upsilon}
%}
%\subfigure[(e) belief]{
%\label{fig:Messages:belief}
%\includegraphics[scale = .7]{fig/belief}
%}
%\MyCaption{Vector-valued messages.}
%\label{fig:Messages}
%\end{figure}

\begin{figure} [t]
\centering
\subfigure[(a) $\rho_{i\rightarrow k}$]{
\label{fig:Messages:rho}
\includegraphics[scale = 1.4]{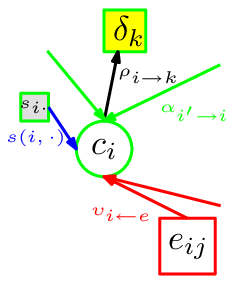}
}~~~~~~~~
\subfigure[(b) $\alpha_{i\leftarrow k}$]{
\label{fig:Messages:alpha}
\includegraphics[scale = 1.4]{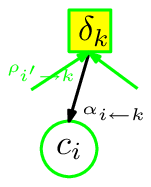}
}~~~~~~~~
\subfigure[(c) $\pi_{i\rightarrow e}$]{
\label{fig:Messages:pi}
\includegraphics[scale = 1.4]{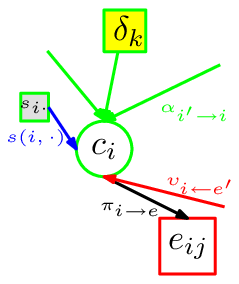}
}~~~~~~~~
\subfigure[(d) $\upsilon_{i\leftarrow e}$]{
\label{fig:Messages:upsilon}
\includegraphics[scale = 1.4]{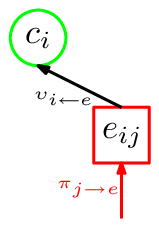}
}~~~~~~~~
\subfigure[(e) belief]{
\label{fig:Messages:belief}
\includegraphics[scale = 1.4]{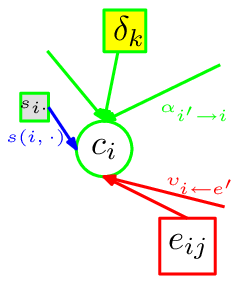}
}
\MyCaption{Vector-valued messages.}
\label{fig:Messages}
\end{figure}

This section presents our proposed
\emph{scalar} message propagation algorithm,
homogeneous and heterogeneous message propagation ($\text{H}^\text{2}\text{MP}$),
which transmits hybrid messages,
over image and tag nodes,
to maximize the objective function~\Equation{\ref{eqn:OverallObjectiveFunction}}.
This algorithm starts from the~\emph{max-sum scheme},
and transform the~\emph{vector} massage propagation
to~\emph{scalar} message propagation
so that the algorithm is very fast.

\mytextbf{\vspace{-.1cm}Naive vector message propagation}
We first present the naive vector message propagation algorithm.
For simplicity, we only give the messages
on the image side,
and the messages on the tag side are similar.
There are two vector messages between $c_i$ and $\delta_k$,
and additionally another message from the heterogeneous relation node $e_{ij}$.
The two vector messages are depicted in~\Figures{\ref{fig:Messages:rho}}{\ref{fig:Messages:alpha}},
and formulated as follows,

{
\vspace{-.2cm}
\footnotesize
\begin{equation}
\rho_{i \rightarrow k}(c_i)
= {\color{red}\sum_{e \in \mathcal{E}^R_i} \upsilon_{i\leftarrow e}(c_i)}
+ s(i, c_i) + \sum_{k': k'\neq k } \alpha_{i\leftarrow k'}(c_i),
\end{equation}
\begin{align}
\alpha_{i \leftarrow k} (c_i)
=& \max_{h_1, \cdots, h_{i-1}, h_{i+1}, \cdots, h_n}[
\sum\nolimits_{i': i' \neq i} \rho_{i' \rightarrow k} (h_{i'}) \nonumber \\
&+ \delta_k(h_1, \cdots, h_{i-1}, c_i, h_{i+1}, \cdots, h_n)].
\end{align}
\vspace{-.2cm}
}

Different from affinity propagation,
we have additional two vector-valued messages
exchanged between $c_i$ and $e_{ij}$.
The message,
$\pi_{i\rightarrow e}$,
sent from variable $c_i$ to $e_{ij}$,
consists of n real numbers,
with one for each possible value of $c_i$.
The message,
$\upsilon_{i\leftarrow e}$,
sent from variable $e_{ij}$ to $c_i$,
also consists of $n$ real numbers.
The two messages are depicted
in~\Figures{\ref{fig:Messages:pi}}{\ref{fig:Messages:upsilon}},
and formulated as follows,

{
\footnotesize
\vspace{-.2cm}
\begin{equation}
\pi_{i \rightarrow e} (c_i)
= s(i, c_i) + \sum_{k}\alpha_{i \leftarrow k} (c_i) + \sum_{e' \in \mathcal{E}^R_i/\{e\}} \upsilon_{i \leftarrow e'} (c_i),
\end{equation}
\begin{equation}
\upsilon_{i \leftarrow e} (c_i) = \max\nolimits_{b_j} [e(c_i, b_j) + \pi_{j \rightarrow e} (b_j)].
\end{equation}
\vspace{-.2cm}
}

%\begin{figure} [t]
%\centering
%\subfigure[(a) $r(i, k)$]{
%\label{fig:ScaleMessages:r}
%\includegraphics[scale = .7]{fig/r1}
%}~~~
%\subfigure[(b) $a(i, k)$]{
%\label{fig:ScaleMessages:a}
%\includegraphics[scale = .7]{fig/a1}
%}~~~
%\subfigure[(c) $w(i, j)$]{
%\label{fig:ScaleMessages:w}
%\includegraphics[scale = .7]{fig/w1}
%}~~~
%\subfigure[(d) $v(i, j)$]{
%\label{fig:ScaleMessages:v}
%\includegraphics[scale = .7]{fig/v1}
%}
%\MyCaption{Scalar-valued messages.}
%\label{fig:ScaleMessages}
%\end{figure}

\begin{figure} [t]
\centering
\subfigure[(a) $r(i, k)$]{
\label{fig:ScaleMessages:r}
\includegraphics[scale = 1.4]{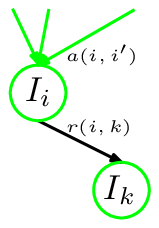}
}~~~~~~~~
\subfigure[(b) $a(i, k)$]{
\label{fig:ScaleMessages:a}
\includegraphics[scale = 1.4]{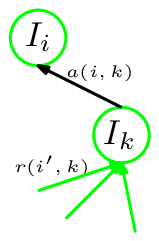}
}~~~~~~~~
\subfigure[(c) $w(i, j)$]{
\label{fig:ScaleMessages:w}
\includegraphics[scale = 1.4]{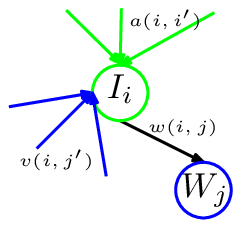}
}~~~~~~~~
\subfigure[(d) $v(i, j)$]{
\label{fig:ScaleMessages:v}
\includegraphics[scale = 1.4]{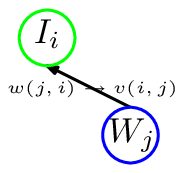}
}
\MyCaption{Scalar-valued messages.}
\label{fig:ScaleMessages}
\end{figure}

One core of this paper is to reduce the above~\emph{vector-valued} messages
to~\emph{scalar}-valued messages.
The derivation is generalized from~\cite{FreyD07},
but it is nontrivial and more challenging
because our problem involves the heterogeneous relations
that cannot be simplified directly using the derivation~\cite{FreyD07}.
Due to space limitation,
we omit the detail derivation\footnote{Please see the supplementary material if the reviewers are interested in the derivation.}
from vector messages to scalar messages.
As a result,
$\text{H}^\text{2}\text{MP}$ views each image or tag as a node in a network,
and recursively transmits~\emph{scalar-valued} messages
along edges of the network
until a good set of image and tag exemplars emerges.
$\text{H}^\text{2}\text{MP}$ is different
from the original affinity propagation algorithm~\cite{FreyD07}
in that $\text{H}^\text{2}\text{MP}$ transmits not only the homogeneous messages
within images and tags,
including responsibility and availability,
and depicted in~\Figures{\ref{fig:ScaleMessages:r}}{\ref{fig:ScaleMessages:a}},
but also the heterogeneous messages between images and tags,
including discardability and contributability,
and depicted in~\Figures{\ref{fig:ScaleMessages:w}}{\ref{fig:ScaleMessages:v}}.
In the following,
we will present four kinds of messages,
and for convenience,
we would only present the homogeneous messages over images
as the messages over tags are similar
and present the heterogeneous messages
by standing at the image side
as the messages on the tag side
can also be similarly obtained.
For presentation simplicity, we drop the subscripts $I$.

\mytextbf{\vspace{-.1cm}Homogeneous message propagation}
The~\Mytextquotedbl{responsibility} and~\Mytextquotedbl{availability} messages
in $\text{H}^\text{2}\text{MP}$ are updated
as follows,

{\footnotesize
\vspace{-.2cm}
\begin{equation}
r(i, k) = {\color{red}\bar{s}(i, k)} - \max\nolimits_{i': i' \neq k}[{\color{red}\bar{s}(i, i')} + a(i, i')].
\label{eqn:RMessage}
\end{equation}
\begin{equation}
{\color{red}
\bar{s}(i, k) = \left\{
\begin{array}{ll}
\sum_{j \in \mathcal{E}^R_{i.}} v(i, j) + s(i, i) & k = i ,\\
s(i, k)& k \neq i .
\end{array}\right.}
\label{eqn:SbarMessage}
\end{equation}
\begin{equation}
a(i, k) = \left\{
\begin{array}{ll}
\sum\nolimits_{i': i' \neq k} \max(0, r(i', k)) & k = i, \\
\min[0, r(k, k) + \sum\nolimits_{i': i' \neq i, k} \max(0, r(i', k))] & k \neq i.
\end{array}\right.
\label{eqn:AMessage}
\end{equation}
\vspace{-.2cm}
}

The key difference in the two messages
from the original affinity propagation
lies in the responsibility $r(i, j)$,
which involves
the heterogeneous message,
i.e., the~\Mytextquotedbl{contributability} message $v(i, j)$
from tag $j$ to image $i$.
This serves as a bridge
in which the affect from tags
will be transmitted to images.
In the iteration process,
$v(i, j)$ would become relatively larger
when the probability of tag $j$
being an exemplar
becomes larger,
and become smaller
otherwise.
Looking at~\Equation{\ref{eqn:SbarMessage}},
we can observe that
the contributability message takes effect
when $k = i$,
which means that it affects the preference
of image $i$ being an exemplar.
Hence,
the probability of image $i$,
selecting itself as its exemplar,
would be affected
positively monotonically
by the probability that tags linking to image $i$
serve as exemplars.

\begin{table}[t]
\MyCaption{
\small{Comparison of the related methods on exploiting the homogeneous (\Mytextquotedbl{homo})
and heterogeneous relations (\Mytextquotedbl{hetero}).
\Mytextquotedbl{simultaneous} means that the two relations
are simultaneously exploited,
and~\Mytextquotedbl{sequential} means that the two relations are sequentially exploited.
Among these approaches,
joint clustering~\cite{RaguramL08} is very close to our approach,
but exploits the heterogeneous relations in a sequential way.
}}
\label{tab:comparisonofmethods}
\begin{center}
\footnotesize{
\begin{tabular}{|@{~}c@{~}||@{~}c@{~}|@{~}c@{~}|@{~}c@{~}|@{~}c@{~}|@{~}c@{~}|}
\hline
& $\text{H}^\text{2}\text{MP}$ & AP/k-means~\cite{SimonSS07} & BGP & TGP & Joint~\cite{RaguramL08}\\
\hline
Homo & Y & Y & N & partial & Y\\
\hline
Hetero & simultaneous & N & Y & Y & sequential \\
\hline
\end{tabular}
}
\end{center}
\end{table}

\mytextbf{\vspace{-.1cm}Heterogeneous message propagation}
There are two kinds of message
exchanged between images and tags.
The~\Mytextquotedbl{discardability}
$w(i, j)$,
sent from image $i$ to tag $j$,
which reflects
how much it is affected that image $i$
selects itself as its exemplar
when the contribution of word $j$ is discarded
and helps tag $j$ make better decision
whether to select itself as its exemplar.
The~\Mytextquotedbl{contributability}
$v(i, j)$,
sent from tag $j$ to image $i$,
which reflects how well image $i$ serves as an exemplar
considering whether tag $j$ is an exemplar.
The two messages are updated as

{
\footnotesize
\vspace{-.2cm}
\begin{align}
w(i, j) = r(i, i) + a(i, i) - v(i, j) = t(i, i) - v(i, j).
\label{eqn:WMessage}
\end{align}
\vspace{-.5cm}
\begin{align}
v(i, j)
=& \max \{ p(i, j), q(i, j) + w(j, i)\} \nonumber \\
&- \max \{\bar{q}(i,j), p(j, i) + w(j, i) \}.
\label{eqn:VMessage}
\end{align}
\vspace{-.2cm}
}

\noindent Here, in~\Equation{\ref{eqn:WMessage}},
$r(i, i) + a(i, i) = t(i, i)$
is the belief that image $i$ selects itself as its exemplar,
and $w(i ,j)$ aims to evaluate
the affect degree if the contribution from tag $j$ to image $i$
is discarded
and help tag $j$ make better decision
whether to select itself as its exemplar.
In evaluating the contributability message $v(i, j)$
from tag $j$ to image $i$
in~\Equation{\ref{eqn:VMessage}},
$w(j, i)$
means that the belief that tag $j$ selects itself as its exemplar
without considering the contribution from image $i$,
and $q(i, j) + w(j, i)$ evaluates the contribution from tag $j$
to the probability that image $i$ serves as an exemplar.
$\max \{ p(i, j), q(i, j) + w(j, i)\}$
essentially means that the degree
that image $i$ serves as an exemplar
whether tag $j$ serves as an exemplar.
Similarly,
$\max \{\bar{q}(i,j), p(j, i) + w(j, i) \}$
means that the degree
that image $i$ does not serve as an exemplar
whether tag $j$ serves as an exemplar.
Their difference,
called contributability,
hence can evaluate how well
image $i$ serves as an exemplar
considering the contribution from tag $j$.
$v(i, j) > 0$
means positive contribution from tag $j$,
and negative contribution otherwise.

\mytextbf{\vspace{-.1cm}Exemplar assignment}
The belief
that image $i$ selects image $j$ as its exemplar
is derived as
the sum of the incoming messages,

{\footnotesize
\vspace{-.2cm}
\begin{equation}
t(i, j) =r(i, j) + a(i, j).
\label{eqn:Belief}
\end{equation}
\vspace{-.2cm}
}

\noindent Then the exemplar of image $i$
is taken as

{\footnotesize
\vspace{-.2cm}
\begin{equation}
\hat{c}_i = \arg\max\nolimits_{j \in \mathcal{E}^I_i\cup\{i\}} t(i, j).
\label{eqn:ExemplarAssignment}
\vspace{-.1cm}
\end{equation}
\vspace{-.2cm}
}

\noindent
It should be noted that
the heterogeneous relations are latently involved in assigning the exemplars
because the responsibility $r(i, j)$ already
counts the contribution from tags
that is indicated in~\Equation{\ref{eqn:RMessage}} and~\Equation{\ref{eqn:SbarMessage}}.

To summarize,
$\text{H}^\text{2}\text{MP}$ is an iterative algorithm,
and at the beginning
all the eight kinds of messages are initialized as 0,
and the eight messages are repeatedly updated
until the iteration number reaches $T$ or the identified exemplars do not change.
The algorithm is described in the following,

\floatname{algorithm}{\small{Algorithm}}
\begin{algorithm}
\algsetup{
   linenosize=\small,
   linenodelimiter=.
}
\MyCaption{Hybrid summarization}\label{algorithm:hap}
\begin{algorithmic}[1]
\small{
    \STATE Initialize all the 8 messages as 0.

    \STATE Compute 4 homogeneous messages for images and tags according to~\Equation{\ref{eqn:RMessage}} and \Equation{\ref{eqn:AMessage}}.

    \STATE Compute 4 heterogeneous messages between images and tags according to~\Equation{\ref{eqn:WMessage}} and \Equation{\ref{eqn:VMessage}}.

    \STATE Repeat steps 2 and 3 till the iteration number reaches $T$ or the identified exemplars do not change.

    \STATE Make image exemplar assignments according to~\Equation{\ref{eqn:ExemplarAssignment}}.
    }
\end{algorithmic}
\end{algorithm}

\subsection{Analysis and discussion}
This subsection presents the time complexity analysis
and discusses the relations of our approach with several existing approaches.

The naive implementation of $\text{H}^\text{2}\text{MP}$
would take $O(T'(n^3 + m^3 + mn(m + n)))$
with $T'$ the iteration number.
Through the trick of reusing some computations,
(e.g., $\max_{i': i' \neq i}[\bar{s}(i, i') + a(i, i')]$ in~\Equation{\ref{eqn:RMessage}},
$\sum_{j \in \mathcal{E}^R_{i.}} v(i, j)$ in~\Equation{\ref{eqn:SbarMessage}},
and $\sum_{i'}\max[0, r(i', k)])$ in~\Equation{\ref{eqn:AMessage}},
are just computed one time for each $i$ in one iteration),
the time complexity of our implementation
is reduced to $O(T'(|\mathcal{E}^I| + |\mathcal{E}^W| + |\mathcal{E}^R|))$
with $|\cdot|$ being the edge number.

Our solution
to hybrid image summarization
is different from
two previous representative techniques
by Simon et al.~\cite{SimonSS07}
and Raguram and Lazebnik~\cite{RaguramL08}
in the following aspects.
Simon et al.
compute the visual summary
by greedy k-means
only using the visual information,
without exploring the useful associated textual information.
Raguram and Lazebnik use a joint clustering method,
which first obtains two independent clusters
from visual and textual features, respectively,
and then takes their intersection to get the final clustering,
but the joint process is obtained sequentially
instead of simultaneously as our approach.

The proposed approach,
$\text{H}^\text{2}\text{MP}$,
is capable to exploit both the relations within images and tags
and the relations between images and tags.
Most related approaches
are only able to capture partial relations.
For example,
AP (affinity propagation~\cite{FreyD07})
can only exploit homogeneous relations over images and words respectively,
BGP (Bipartite graph partitioning~\cite{Dhillon01})
can only exploit heterogeneous relations
between images and tags,
TGB (Tripartite graph partitioning~\cite{RegeDH08})
uses the visual features
besides heterogeneous relations
between images and tags
to help find the grouping
without exploring the interior relations within tags.
The comparison is summarized in~\Table{\ref{tab:comparisonofmethods}}.

\subsection{Implementation}
Similar to AP~\cite{FreyD07},
the self-similarity $s'_I(i, i)$ of an image $i$,
i.e., the preference of an image being an exemplar,
is set as $\lambda\operatorname{Med}[s'_I(i, k)]$
with $\operatorname{Med}[s'_I(i, k)]$
being the median image similarity.
$\lambda$ is useful
to control the exemplar number.
For tags,
we adopt the WordNet similarity~\cite{WordNetSimilarity},
a variety of semantic similarity and relatedness measures
based on a large lexical database of English,
WordNet~\cite{WordNet}.
The self-similarities of words
are similarly set.

Let's turn to the setting of $\gamma_I$ and $\gamma_W$
in~\Equation{\ref{eqn:OverallObjectiveFunction}}.
Looking at~\Equation{\ref{eqn:SbarMessage}},
we observed that
the heterogeneous relations
essentially adjust the preference of image $i$,
$\bar{s}(i, i)$,
through the contributability $v(i, j)$ from tag $j$ to $i$,
and hence it is expected
that $v(i, j)$ is comparable with the preference $s(i, i)$.
Furthermore,
we observed that
$v(i, j)$ is computed from $w(j, i)$
in~\Equation{\ref{eqn:VMessage}}
and $w(j, i)$ sent from tag $j$ to image $i$
is computed from the belief $t(j, j)$ of word $j$
being an exemplar
that is related to tag similarities
in~\Equation{\ref{eqn:WMessage}}.
Thus,
to make $p(\cdot, \cdot)$ in the heterogeneous relations
easily tuned, which may benefit
from the comparable preferences of tag and image,
in our experiment we fix $\gamma_I = \frac{1}{\operatorname{Med}(s'_I(i, k))}$
and $\gamma_W = \frac{1}{\operatorname{Med}(s'_W(j, k))}$.

For $p(\cdot, \cdot)$ in the heterogeneous relations $e_{ij}(c_i, b_j)$,
we set $p(i,j)=\theta/|\mathcal{E}^R_{i.}|$,
and $p(j,i)=\theta/|\mathcal{E}^R_{.j}|$,
where $\theta$  is a constant negative value,
fixed as 15 in this paper,
to control the mutual affect degree for image and tag exemplar identification,
and the division by the tag number connecting image $i$,
$|\mathcal{E}^R_{i.}|$, aims to averagely separate its affect to connected tags.

\section{Experiments}
\label{sec:experiment}
In our experiment,
we present the performance comparison of our approach
with several relate approaches.
This collection consists of about 11k images and associated tags
and is crawled
from the popular photo sharing Web site Flickr,
using the queries,
including flower, city, building, dog, cat, plants, mountain, river, sunset, and so on.
We filter out some noisy tags
that few images are associated with
and finally get 816 tags.
On average,
each image has 6.1 tags
and each tag is assigned to 15.9 images.
We extract a GIST scene descriptor~\cite{OlivaT01},
which has been shown to work well for scene categorization,
as the image feature
with
3 by 3 spatial resolution where each bin contains that image
region's average response to steerable filters at 6 orientations
and 3 scales,
and use the negative Euclidean distance as the image similarity.

We investigate the performances
from both the visual and semantic perspectives.
In the literature of image summarization and clustering,
most evaluation criteria use the class labels
of the images to test the performance.
However, they are not adoptable for our hybrid summarization
because hybrid summarization has multiple objectives,
visual and sematic objectives
and no simple labels can be applied here.
Instead,
we present two straightforward measures,
visual exemplarness and sematic exemplarness.
Visual exemplarness is defined
as the average value
of visual similarities between each image and its corresponding exemplar,
and semantic exemplarness is defined
as the average value
of textual similarities
between the associated tags of each image and its corresponding exemplar.

\begin{figure}[t]
\centering
\subfigure[(a)]
{\includegraphics[width = .38\linewidth]
{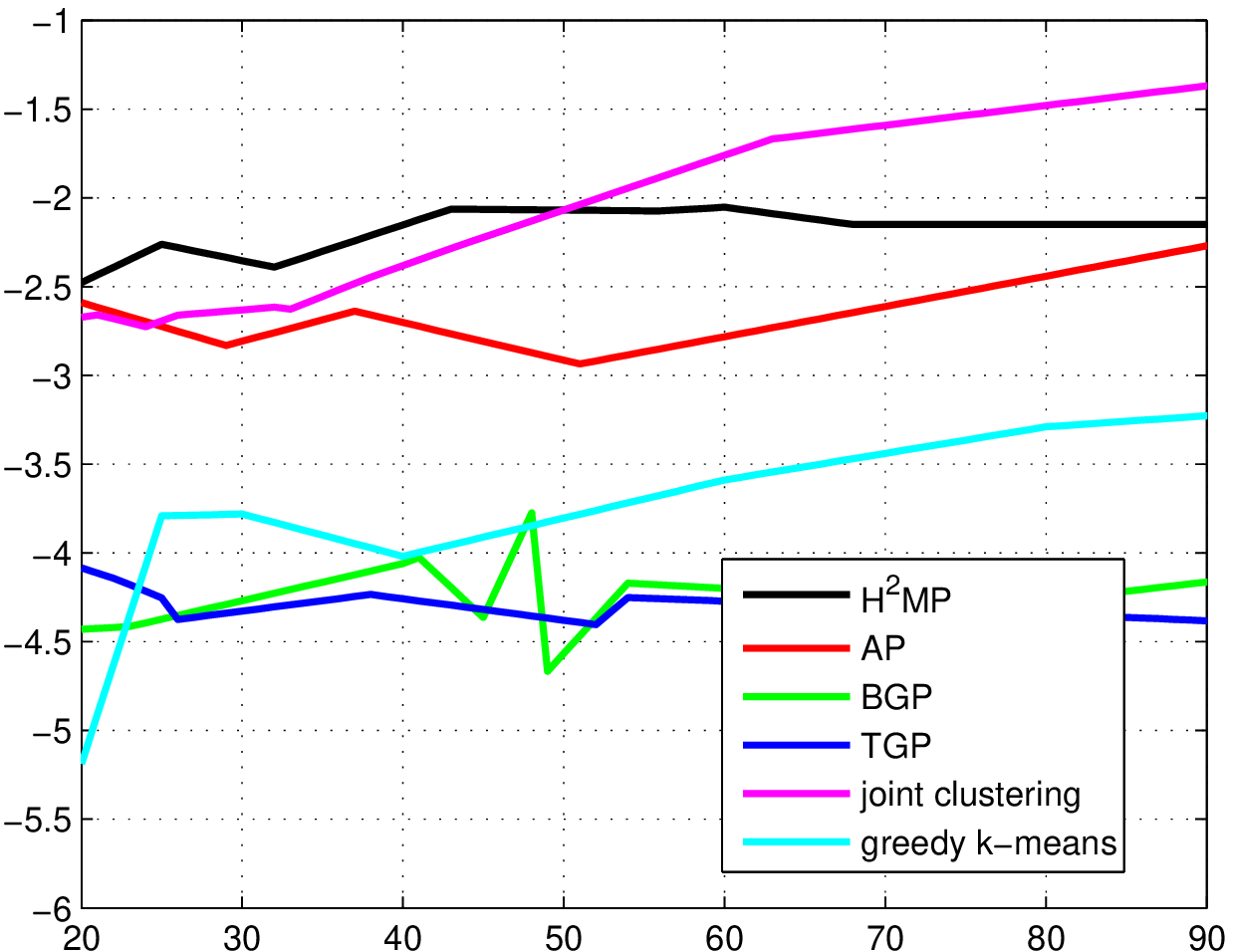}}~~~~~~
\subfigure[(b)]
{\includegraphics[width = .38\linewidth]
{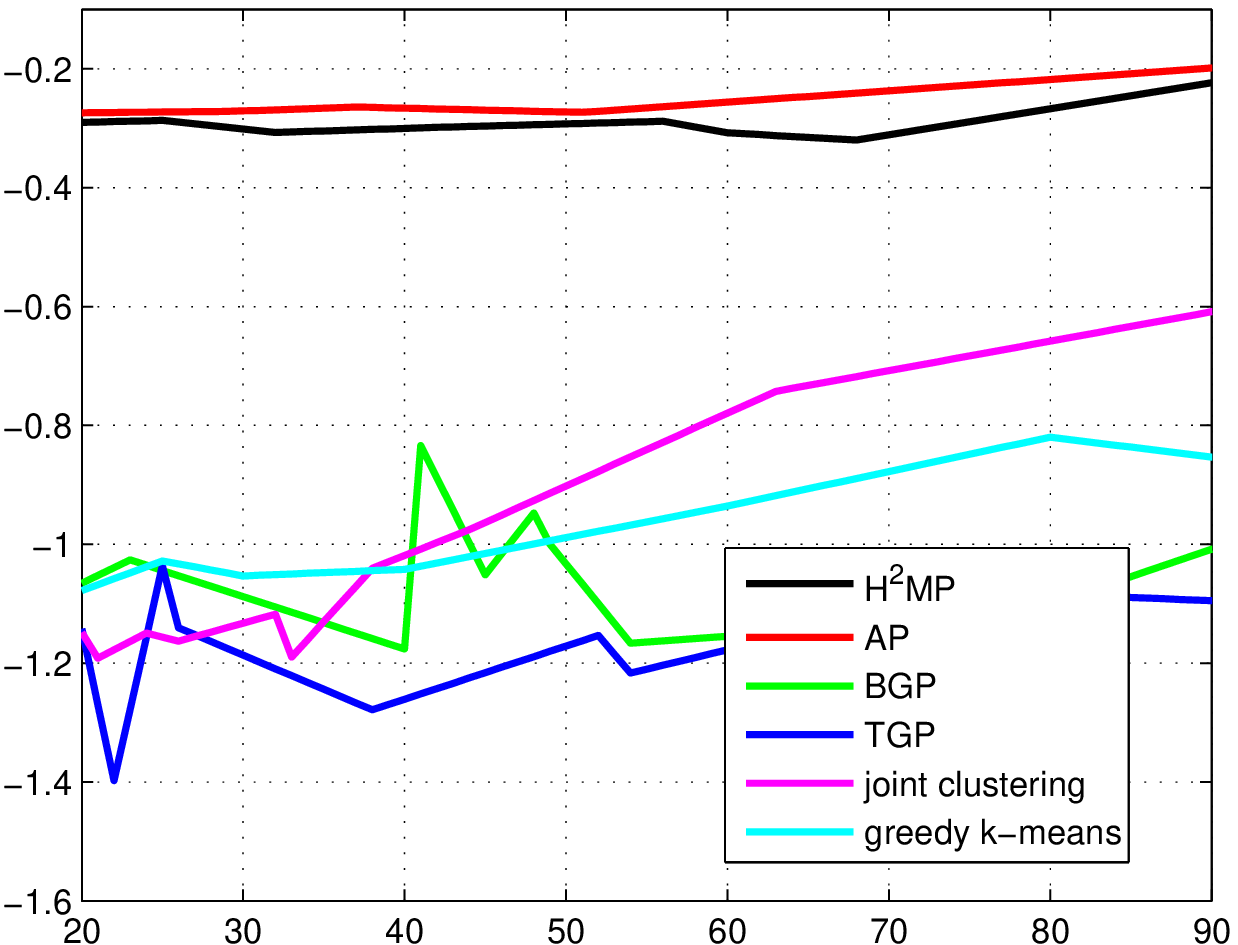}}
\MyCaption
{Performance comparison with related approaches.
The x-axis represents the exemplar number,
the y-axis in (a) and (b) represent the semantic exemplarness
and visual exemplarness.
}
\label{fig:PerformanceComparison}
\end{figure}

\subsection{Quantitative comparison}
We present a quantitative comparison
of our approach ($\text{H}^\text{2}\text{MP}$)
with several representative approaches,
AP (affinity propagation~\cite{FreyD07}),
BGP (Bipartite graph partitioning~\cite{Dhillon01}),
TGB (Tripartite graph partitioning~\cite{RegeDH08}),
and recently developed two methods:
greedy k-means~\cite{SimonSS07} and
joint clustering~\cite{RaguramL08}.
\Figure{\ref{fig:PerformanceComparison}(a)} and~\Figure{\ref{fig:PerformanceComparison}(b)}
illustrate the performances of different approaches
in terms of semantic and visual exemplarness
with different number of exemplars.

For semantic exemplarness,
$\text{H}^\text{2}\text{MP}$ constantly outperforms the other approaches
except the joint clustering approach~\cite{RaguramL08}
and its performance is a little worse
than the joint clustering approach
when the number of exemplars exceeds 50.
This is understandable
because our approach balances
the visual and semantic performances
while joint clustering generates
results by taking intersection
between the results
using visual feature
and text feature to cluster images,
and hence may get superiority for semantic performance
when the cluster number is very large.
However,
the performance for modest number of exemplars is more meaningful,
because too many exemplars are not preferred in summarization.
From this sense,
our approach is more satisfactory in semantic performance.

For visual exemplarness,
both AP and $\text{H}^\text{2}\text{MP}$ show significant advantages over the other approaches.
The visual performance of $\text{H}^\text{2}\text{MP}$ is only a little worse
than that of AP that purely uses visual feature,
which is reasonable
since our approach also takes into consideration
the semantic information.
In summary,
$\text{H}^\text{2}\text{MP}$ achieves satisfactory semantic and visual performance
compared with other approaches.

\begin{table}[t]
\centering
\MyCaption
{Quantitative results
on three groups:
\Mytextquotedbl{Anchorage, Alaska},~\Mytextquotedbl{Roma-Rome}
and~\Mytextquotedbl{The Great Wall of China}.
\Mytextquotedbl{S} represents the semantic exemplarness,
and \Mytextquotedbl{V} denotes the visual exemplarness.
The images in~\Mytextquotedbl{The Great Wall of China}
have few tags
and hence the semantic exemplarness measure can not be evaluated.
The best two scores are highlighted
in bold font.}
\label{tab:GroupQuantatitveComparison}
\centering
{\small
\begin{tabular}{|@{~}c@{~}||@{~~}c@{~~}|@{~~}c@{~~}||@{~~} c@{~~} |@{~~} c@{~~} || @{~~}c@{~~}|@{~~}c@{~~} |}
\hline
& S & V & S & V &S & V\\
\hline
\hline
$\text{H}^\text{2}\text{MP}$ & \textbf{-2.264} & {-0.761} & \textbf{-0.901} & -0.775 & -& \textbf{-0.399} \\
\hline
AP & -3.225 &  \textbf{-0.745} & -2.473 & \textbf{-0.706}  & -& \textbf{-0.396}\\
\hline
BGP &  -4.161 & -0.938 &  -4.637 & -0.919  & -& -0.407\\
\hline
TGP & -3.702 &  -0.928 & -3.266 & -0.802  &- & -0.593\\
\hline
Joint~\cite{RaguramL08} & \textbf{-2.205} & -0.917 & -3.596 & -1.062  &- & - 0.499\\
\hline
k-means~\cite{SimonSS07} & -2.283 & \textbf{-0.703} & \textbf{-2.202} & \textbf{-0.650}  & -& -0.412\\
\hline
\end{tabular}
}
\end{table}

\subsection{Visual comparison}
We present visual results
on three representative groups of images from Flickr,
\Mytextquotedbl{Anchorage, Alaska},
\Mytextquotedbl{Roma-Rome},
and~\Mytextquotedbl{The Great Wall of China}.
We crawled top 970, 928, and 133 images, respectively.

\begin{figure} [t]
\centering
\begin{tabular}{c@{~~}c@{~~~~~}|@{~~~~~}c}
{\footnotesize (a)}&\includegraphics[width = .4\linewidth, height = .073\linewidth]
{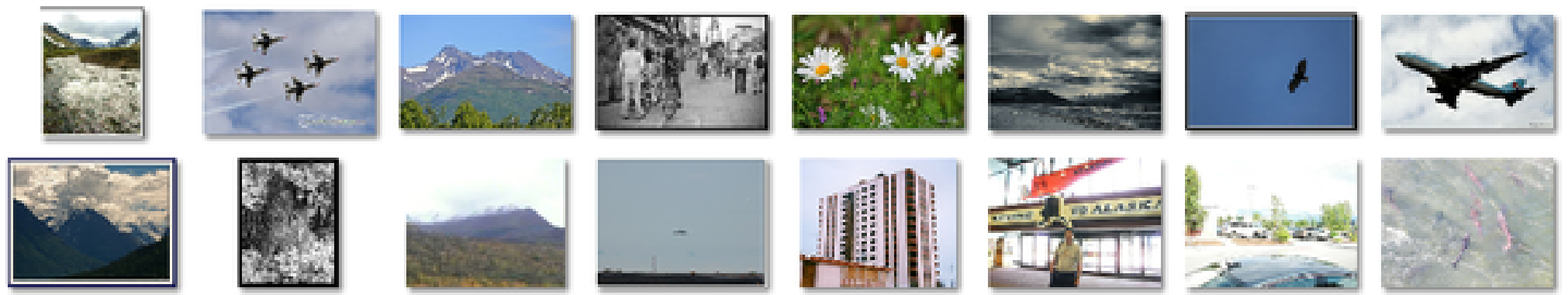}
& \includegraphics[width = .4\linewidth, height = .073\linewidth]
{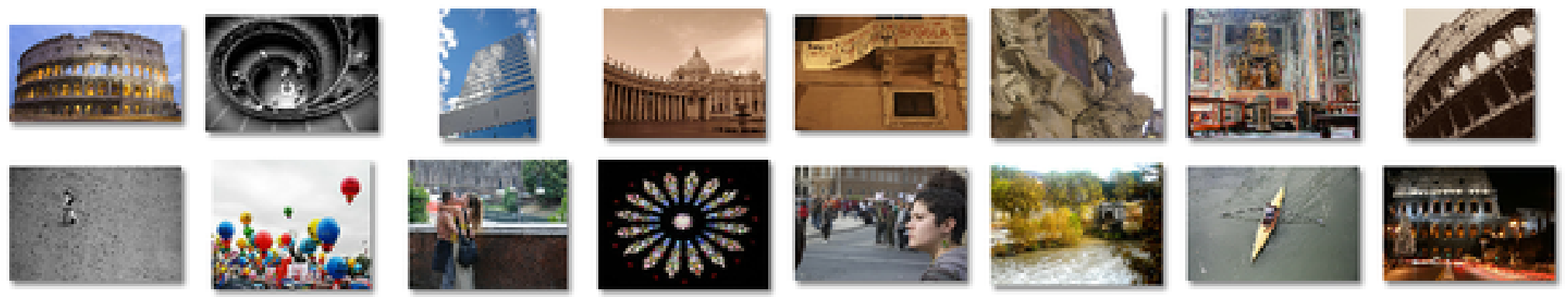}
\\
\hline
{\footnotesize (b)}&\includegraphics[width = .4\linewidth, height = .167\linewidth]
{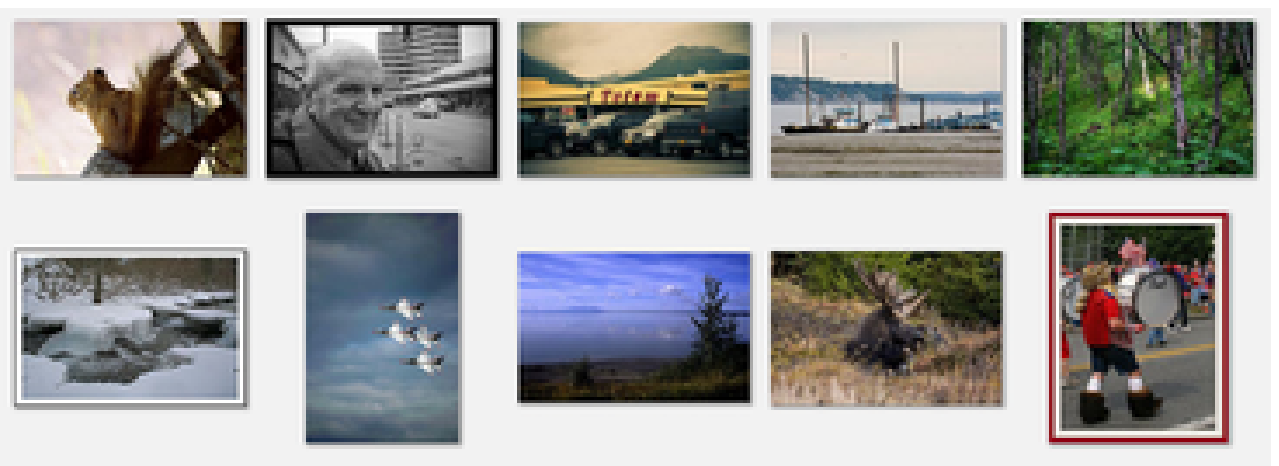}
& \includegraphics[width = .4\linewidth, height = .167\linewidth]
{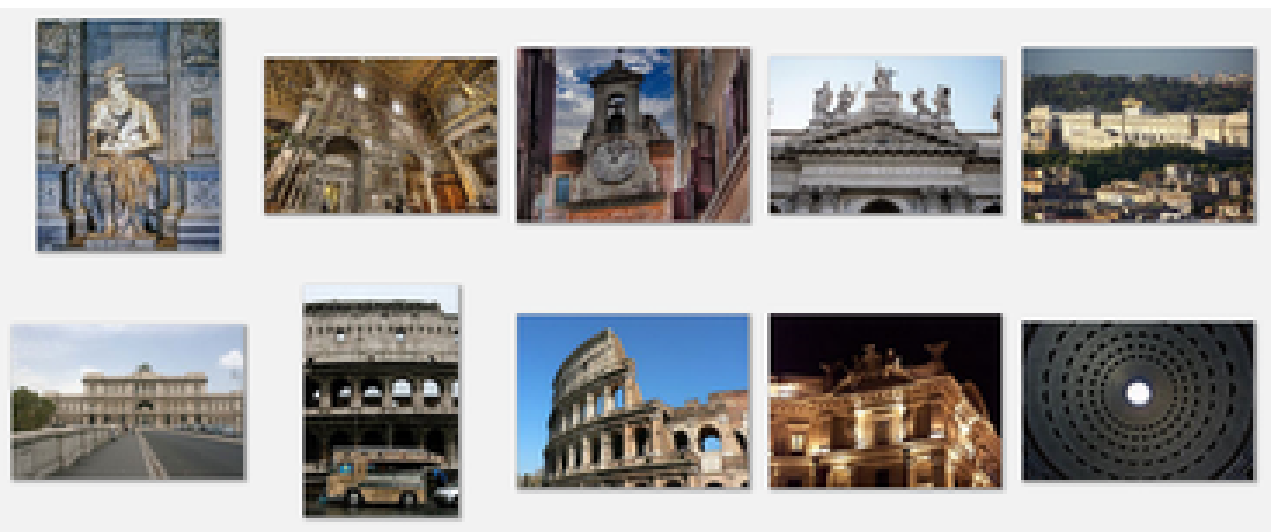}
\\   \hline
{\footnotesize(c)}&\includegraphics[width = .4\linewidth, height = .167\linewidth]
{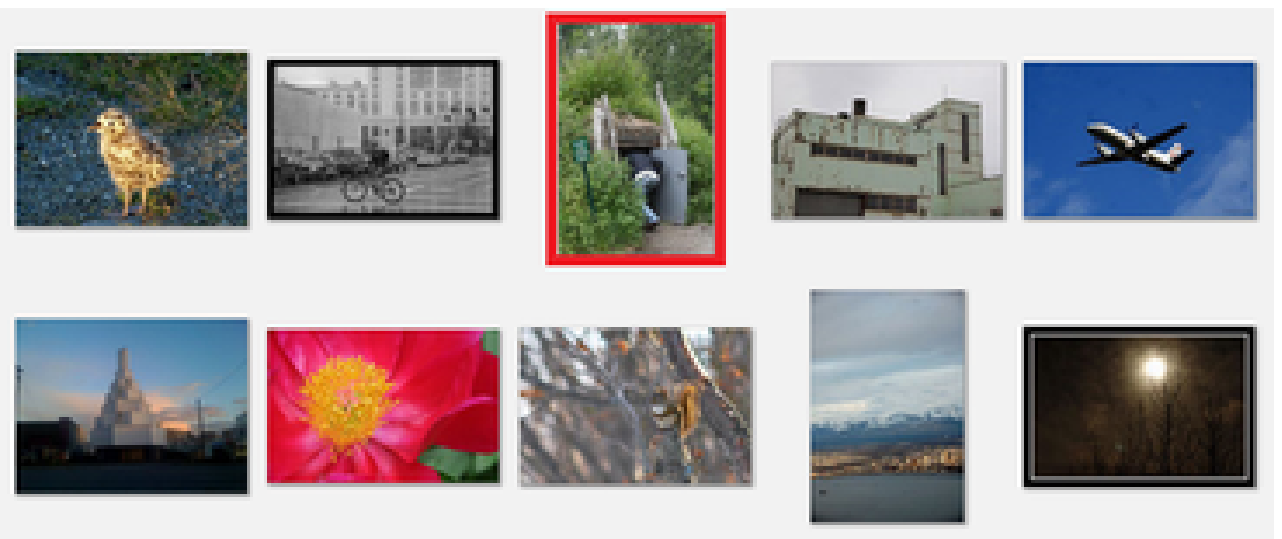}
& \includegraphics[width = .4\linewidth, height = .167\linewidth]
{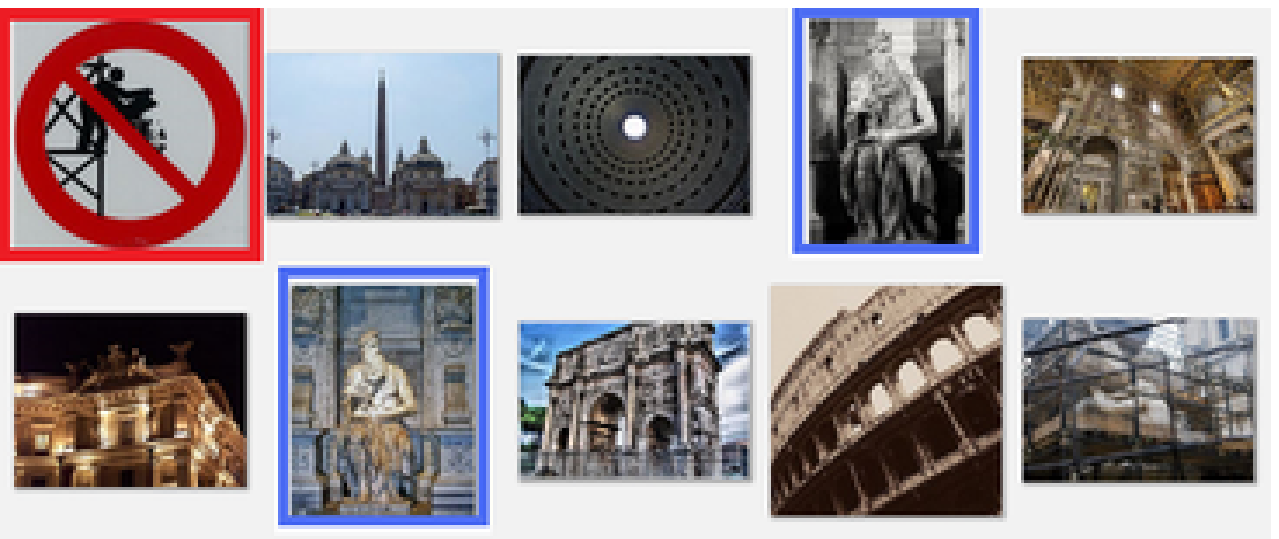}
\\        \hline
{\footnotesize (d)}&\includegraphics[width = .4\linewidth, height = .167\linewidth]
{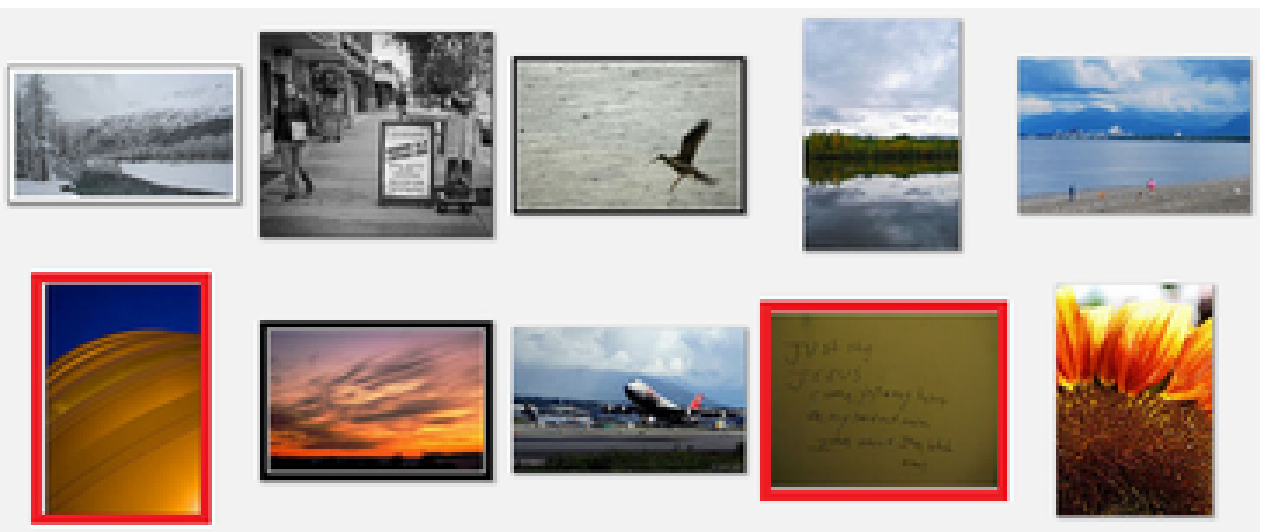}
& \includegraphics[width = .4\linewidth, height = .167\linewidth]
{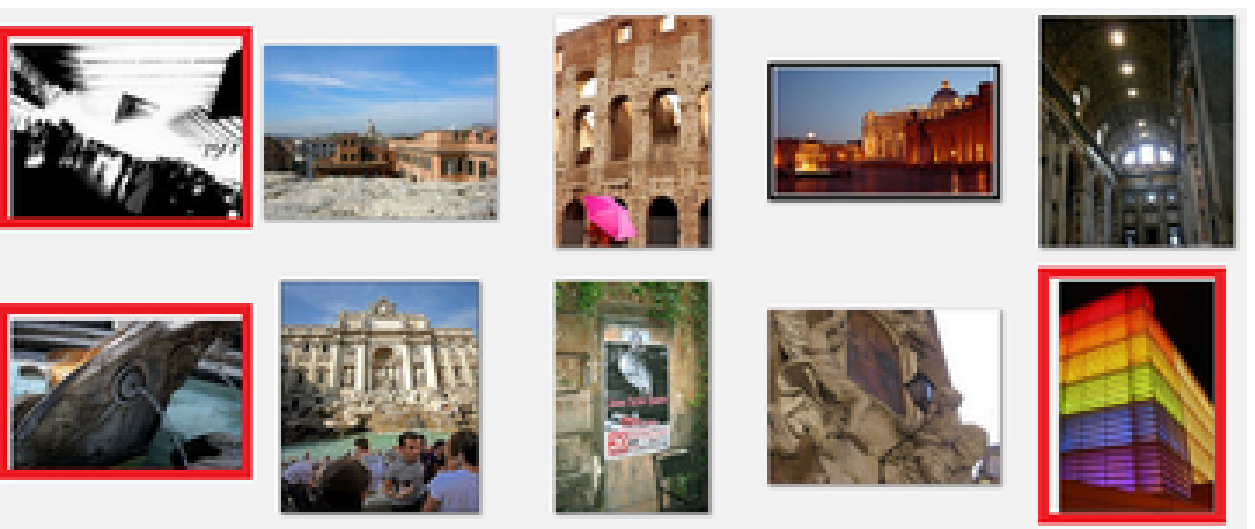}
 \\              \hline
{\footnotesize (e)}&\includegraphics[width = .4\linewidth, height = .167\linewidth]
{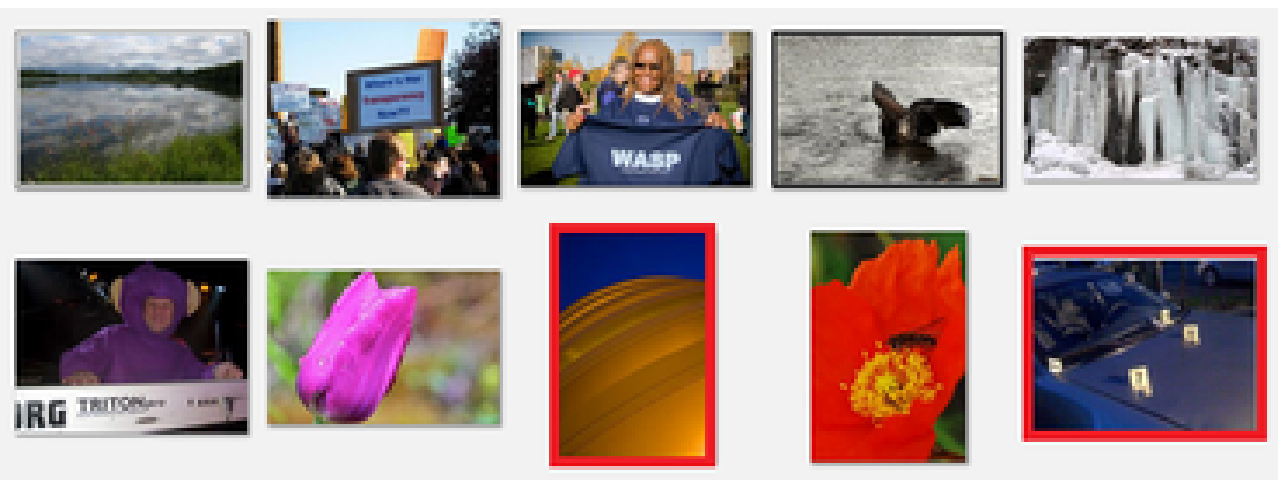}
& \includegraphics[width = .4\linewidth, height = .167\linewidth]
{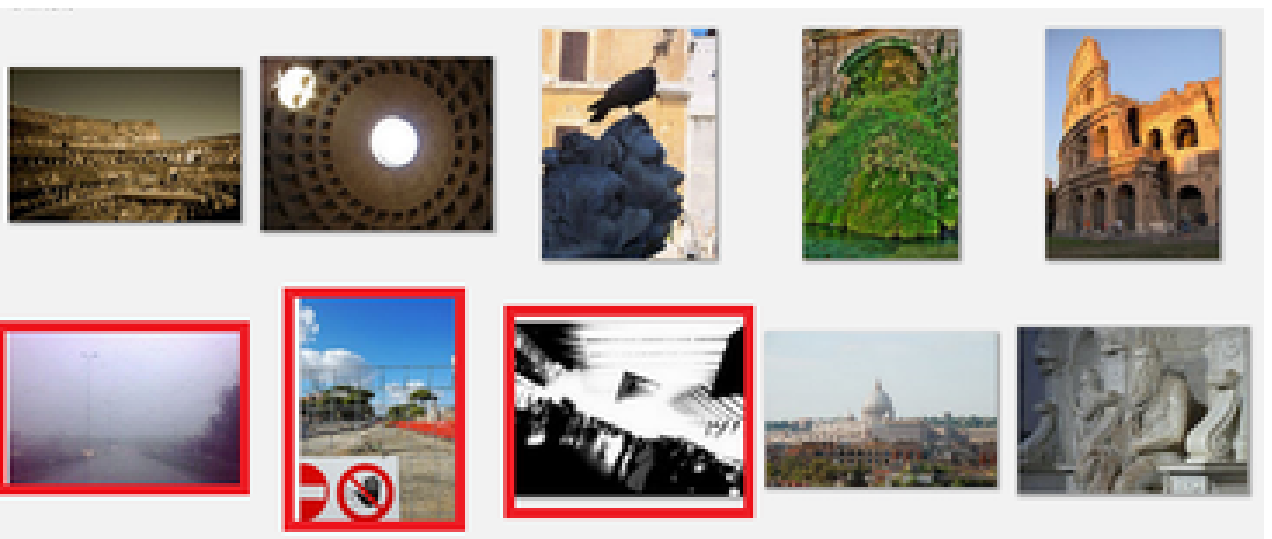}
\\       \hline
{\footnotesize (f)} &\includegraphics[width = .4\linewidth, height = .167\linewidth]
{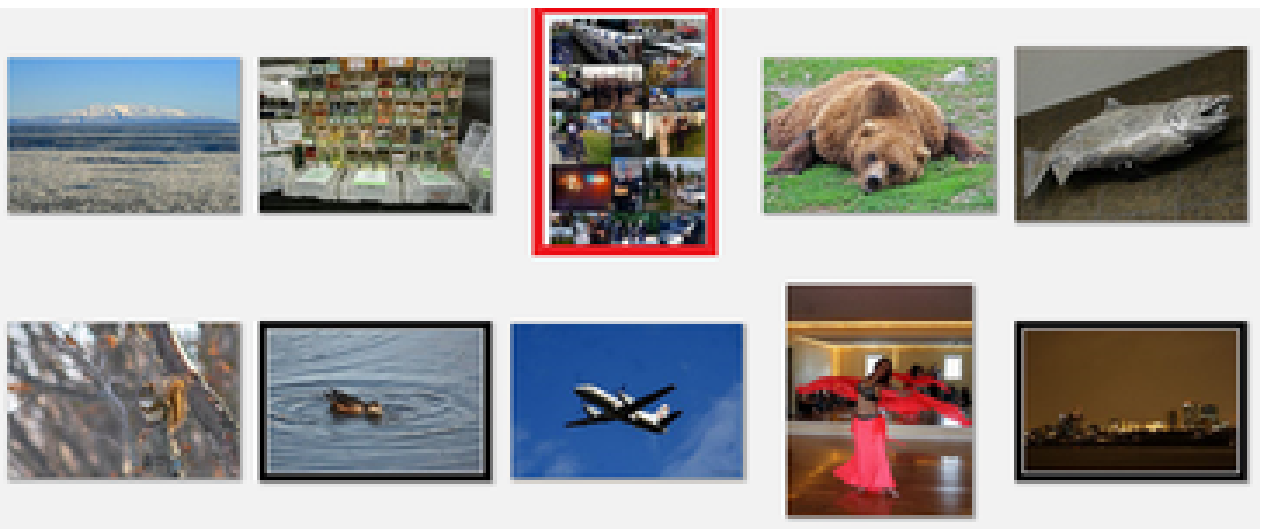}
& \includegraphics[width = .4\linewidth, height = .167\linewidth]
{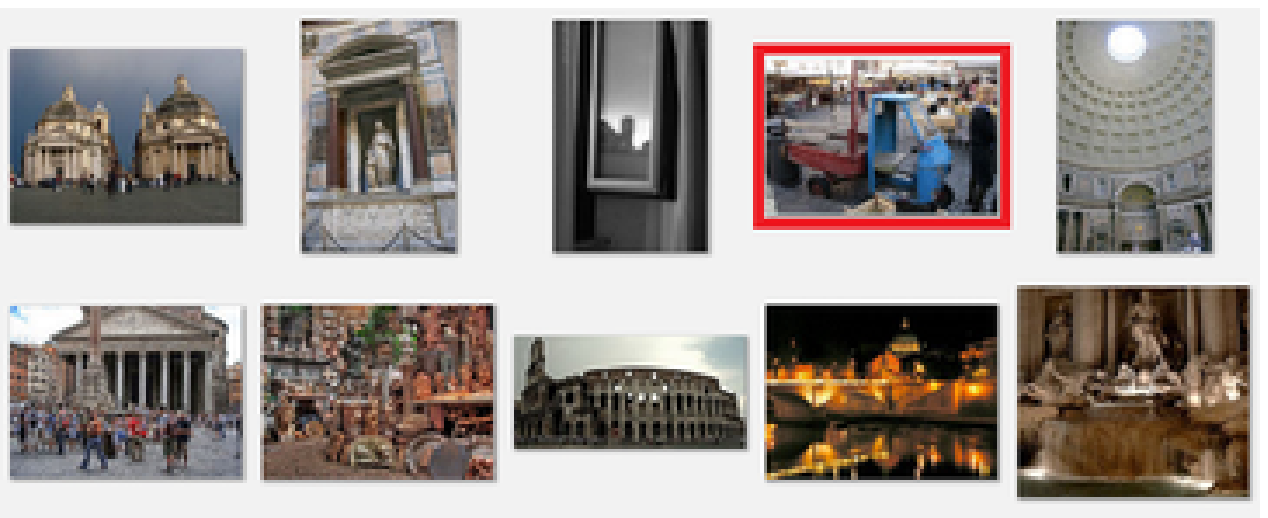}
\\        \hline
{\footnotesize (g)}&\includegraphics[width = .4\linewidth, height = .167\linewidth]
{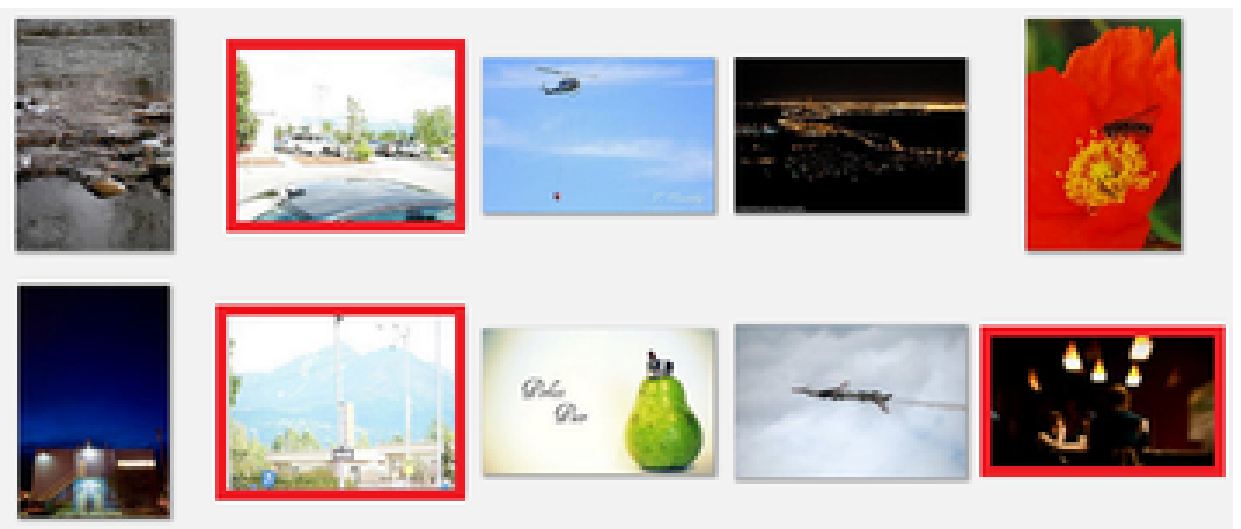}
& \includegraphics[width = .4\linewidth, height = .167\linewidth]
{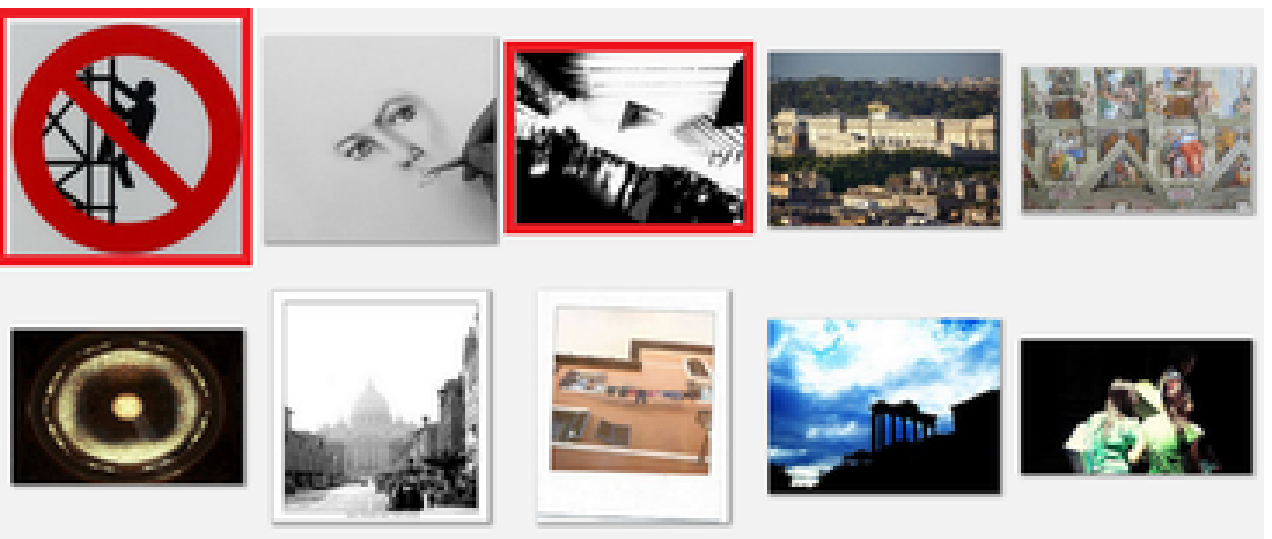}
\end{tabular}
\MyCaption
{Visual results on~\Mytextquotedbl{Anchorage, Alaska} (left column)
and~\Mytextquotedbl{Roma-Rome} (right column).
(a) the examples of original images,
(b)- (g) correspond to the results of
$\text{H}^\text{2}\text{MP}$, AP, BGP, TGP, joint clustering and greedy k-means.}
\label{fig:visualComparison}
\end{figure}

The visual results of~\Mytextquotedbl{Anchorage, Alaska}
\Mytextquotedbl{Roma-Rome}
from the six approaches
are depicted in~\Figure{\ref{fig:visualComparison}},
and their quantitative comparison is in~\Table{\ref{tab:GroupQuantatitveComparison}}.
Our results look visually appealing,
and the superiority in semantic performances shows
that the obtained visual summary can capture the semantic meaning,
which benefits from the associated tags.
The other methods cannot get competitive performance
because those methods have partial or little ability to
exploit homogeneous and heterogeneous relations.

\subsection{User study}
In addition,
we present a user study
to compare visual summaries
of six approaches.
We collect the feedback from 20 persons
on 10 tagged image collections.
For each person,
we show him a tagged image collection
and randomly select a visual summary
from the six results corresponding six approaches,
and allow him to given a score from 1 (the worst) to 5 (the best)
to indicate how well the visual summary represents
the image collection from the visual and semantic perspectives.
The user study shows that
our approach obtains the best performance 4.3,
and the scores of other approaches,
AP, BGP, TGP, joint clustering~\cite{RaguramL08} and greedy k-means~\cite{SimonSS07}
are 3.6, 3, 3.2, 3.9, and 3.8, respectively.
This user study demonstrates
that our approach can get better visual summary
compared with other approaches.

\subsection{Application}
\begin{figure*}[t]
\centering
{
\begin{tabular}{cc}
\multirow{2}{*}[3cm]
{\includegraphics[width = .68\linewidth]{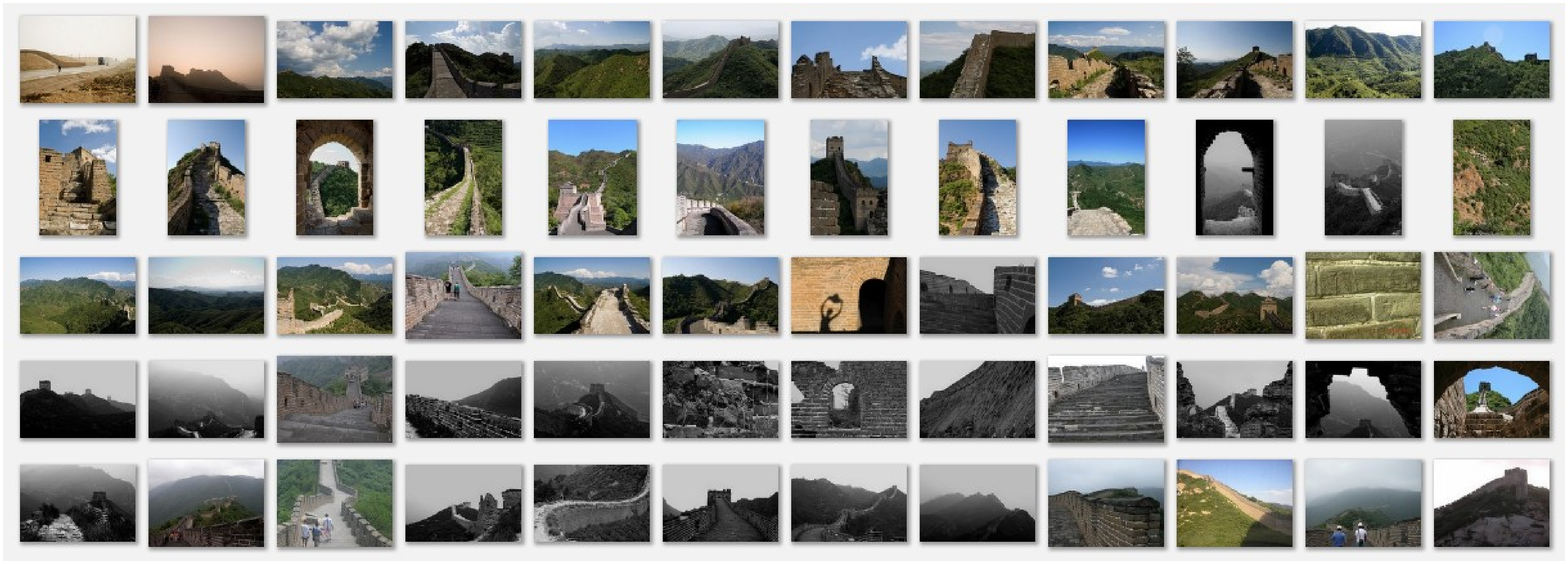}}
& \includegraphics[width = .27\linewidth]{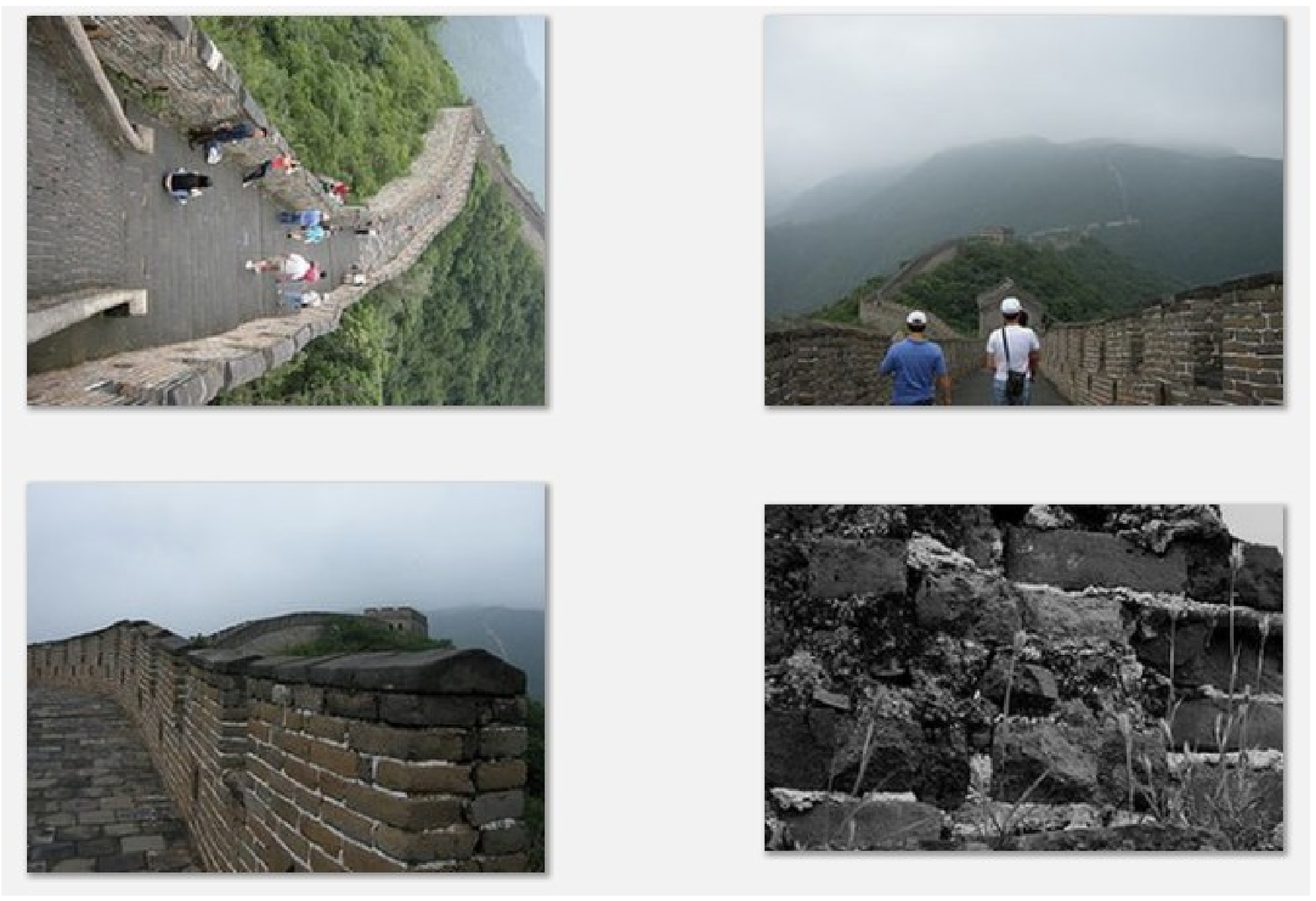} \\
& {\footnotesize (b) Visual summary} \\
& {\footnotesize \color{blue}simatai, mutianyu} \\
{\footnotesize (a) Sample images in the group} & {\footnotesize (c) Textual summary}
\end{tabular}
}
\MyCaption{Flickr group summarization for~\Mytextquotedbl{The Great Wall of China}.}
\label{fig:ImageSearch}
\end{figure*}

We demonstrate the hybrid summarization
in Flickr group overview
by presenting both image and text summarization.
\Mytextquotedbl{Flickr groups are a fabulous way
to share content and conversation,
either privately or with the world.
Believe us when we say there's probably a group for everyone,
but if you can't find one you like,
feel free to start your own.}.
The group images are displayed
page by page,
and each page shows a dozen of images.
To have an overview
of a group of images,
uses have to check the images
page by page,
and also there is no textual description
for the group of images.
Hence it is desired
to deliver visual and textual summaries of the group.
\Figure{\ref{fig:ImageSearch}} shows an example
over~\Mytextquotedbl{The Great Wall of China}.
It is surprisedly that the hybrid summarization suggests two tags:
simatai and mutianyu.
After checking this group manually,
we found that this group only contains the photos from two sites.

\section{Conclusion}
\label{sec:conclusion}
In this paper,
we present hybrid image summarization scheme
to manage image collections.
Toward this end,
we propose a novel approach,
homogeneous and heterogeneous message propagation,
which is a non-trivial generalization of
the affinity propagation algorithm
from homogeneous data
to heterogeneous data.
Compared with the conventional message propagation algorithms
that transmit the vector-valued messages,
our algorithm reduces vector-valued messages
to scalar-valued messages,
and hence is more efficient.
Moreover, this reduction in our case is more complicated
than in affinity propagation
because it involves additional heterogeneous relations.
The application of our approach to hybrid image summarization
can effectively exploit image similarities
and even the useful information from the associated tags,
including the association relations between images and tags
and the relations within tags.
The experimental results
demonstrate its effectiveness and efficiency.

\section*{Appendix}
\subsection*{Derivation}
We rewrite the objective function,
which corresponds to Eqn. {\color{red}(11)} in the submitted paper.
\begin{align}
\mathcal{S}(\mathbf{c}, \mathbf{b})
=& \gamma_I (E^I(\mathbf{c}) + V^I(\mathbf{c}))
+ \gamma_W (E^W(\mathbf{b}) + V^W(\mathbf{b}))
+ R(\mathbf{c}, \mathbf{b}) \nonumber \\
=& \sum\nolimits_{i = 1}^{n} s_I(i, c_i)
+ \sum\nolimits_{k = 1}^{n} \delta_k(\mathbf{c})
+ \sum\nolimits_{j = 1}^{m} s_W(j, b_j)
+ \sum\nolimits_{k = 1}^{m} \eta_k(\mathbf{b})
+ \sum\nolimits_{(i, j) \in \mathcal{E}^R }e_{ij}(c_i, b_j).
\label{eqn:OverallObjectiveFunction}
\end{align}

This objective function is depicted
as a factor graph~\Figure{\ref{fig:FactorGraphForOverallObjective}}.
The max-sum algorithm~\cite{KschischangFL01},
a general algorithm
to factor graph,
can get the solution
by transmitting vector messages between function nodes and variable nodes.
We would derive a scalar message propagation scheme,
homogeneous and heterogeneous message propagation,
by reducing vector messages over function and variable nodes
to scalar messages over variable nodes.

%\begin{figure}[t]
%\centering
%\includegraphics[scale = 1.]{fig/bifactorgraph}
%\MyCaption{Factor graph for
%the overall objective function~\Equation{\ref{eqn:OverallObjectiveFunction}}.
%$\square$ represents a function node,
%and $\bigcirc$ represents a variable node.}
%\label{fig:FactorGraphForOverallObjective}
%\end{figure}

The max-sum algorithm is an iterative algorithm
to exchange two kinds of messages:
one is from function nodes to variable nodes,
and the other is from variable nodes to function nodes.
For the factor graph~\Figure{\ref{fig:FactorGraphForOverallObjective}}
corresponding to~\Equation{\ref{eqn:OverallObjectiveFunction}},
the message propagation over variables $c_i$ and $b_j$
is almost the same,
For convenience we will give the derivation over variable $c_i$,
and drop the subscript $I$ in $s_I(\cdot, \cdot)$ in the following presentation.

There are two messages
exchanged between $c_i$ and $\delta_k$.
The message, $\rho_{i \rightarrow k}$,
sent from $c_i$ to $\delta_k$,
consists of $n$ real numbers,
one for each possible value of $c_i$.
The message,
$\alpha_{i\leftarrow k}$,
sent from $\delta_k$ to $c_i$,
also consists of $n$ real numbers.
The two messages are depicted
in~\Figures{\ref{fig:Messages:rho}}{\ref{fig:Messages:alpha}},
and formulated as follows:

\begin{equation}
\rho_{i \rightarrow k}(c_i)
= \sum_{e \in \mathcal{E}^R_i} \upsilon_{i\leftarrow e}(c_i)
+ s(i, c_i) + \sum_{k': k'\neq k } \alpha_{i\leftarrow k'}(c_i),
\end{equation}

\begin{align}
\alpha_{i \leftarrow k} (c_i)
=& \max_{h_1, \cdots, h_{i-1}, h_{i+1}, \cdots, h_n}[
\sum\nolimits_{i': i' \neq i} \rho_{i' \rightarrow k} (h_{i'}) + \delta_k(h_1, \cdots, h_{i-1}, c_i, h_{i+1}, \cdots, h_n)].
\end{align}

There are two messages
exchanged between $c_i$ and $e_{ij}$.
The message,
$\pi_{i\rightarrow e}$,
sent from variable $c_i$ to $e_{ij}$,
consists of n real numbers,
one for each possible value of $c_i$.
The message,
$\upsilon_{i\leftarrow e}$,
sent from variable $e_{ij}$ to $c_i$,
also consists of $n$ real numbers.
Let $\mathcal{E}^R_i = \{e_{ij}\}_j$
represent the edge set connecting image $i$.
For simplicity,
we drop the subscript ${_ij}$ in $e_{ij}$
without influencing understanding.
The two messages are depicted
in~\Figures{\ref{fig:Messages:pi}}{\ref{fig:Messages:upsilon}},
and formulated as follows:

\begin{equation}
\pi_{i \rightarrow e} (c_i)
= s(i, c_i) + \sum_{k}\alpha_{i \leftarrow k} (c_i) + \sum_{e' \in \mathcal{E}^R_i/\{e\}} \upsilon_{i \leftarrow e'} (c_i),
\end{equation}

\begin{equation}
\upsilon_{i \leftarrow e} (c_i) = \max\nolimits_{b_j} [e(c_i, b_j) + \pi_{j \rightarrow e} (b_j)].
\end{equation}

In the following,
we show that those vector-valued messages
can be reduced to scalar-valued messages,
making the propagation much more efficient.
The derivation is generalized from~\cite{FreyD07},
but it is nontrivial and more challenging
since the message is additionally
propagated between heterogeneous data,
images and words.
We directly present the results
for $\rho-$ and $\alpha-$ messages
by omitting detailed derivation
that can be obtained
using the similar technique as in~\cite{FreyD07}.
We present the derivation detail for
$\upsilon-$ and $\pi-$ messages.
The idea behind the derivation
is to analyze the propagated messages
in the two cases
whether $c_i$ is valued as $i$ or not.

Let $\tilde{\rho}_{i \rightarrow k}(c_i) = \rho_{i \rightarrow k} (c_i) - \bar{\rho}_{i \rightarrow k}$,
with $\bar{\rho}_{i \rightarrow k} = \max_{h: h \neq k} \rho_{i \rightarrow k} (h)$.

Let $\tilde{\alpha}_{i \leftarrow k} (c_i) = \alpha_{i \leftarrow k} (c_i) - \bar{\alpha}_{i \leftarrow k}$,
with $\bar{\alpha}_{i \leftarrow k} = \alpha_{i \leftarrow k} (c_i : c_i \neq k)$.
It can be derived that $\alpha_{i \leftarrow k} (c_i : c_i \neq k)$ is independent to the specific value $c_i$.

Let $\tilde{\upsilon}_{i \leftarrow e}(c_i) = \upsilon_{i \leftarrow e} (c_i) - \bar{\upsilon}_{i \leftarrow e}$,
\begin{align}
\bar{\upsilon}_{i \leftarrow e} &= \upsilon_{i \leftarrow e} (c_i : c_i \neq i) \nonumber \\
& = \max\nolimits_{b_j} [ e(c_i, b_j) + \pi_{j \rightarrow e}(b_j)] \nonumber\\
& = \max [ \max\nolimits_{b_j: b_j \neq j} [ e(c_i, b_j) + \pi_{j \rightarrow e}(b_j) ], e(c_i, j) + \pi_{j \rightarrow e}(j) ] \nonumber\\
& = \max [ \bar{q}(i,j) + \max\nolimits_{b_j: b_j \neq j} \pi_{j \rightarrow e}(b_j), p(j, i) + \pi_{j \rightarrow e}(j) ].
\end{align}

Let $\tilde{\pi}_{i \rightarrow e} (c_i) = \pi_{i \rightarrow e}(c_i) - \bar{\pi}_{i \rightarrow e}(c_i)$,
and
\begin{align}
\bar{\pi}_{i\rightarrow e} &= \max\nolimits_{c_i: c_i \neq i} \pi_{i\rightarrow e} (c_i) \nonumber\\
&= \max_{c_i: c_i \neq i} [
s(i, c_i) + \tilde{\alpha}_{i \leftarrow c_i} (c_i) + \sum_{e' \in \mathcal{E}^R_i/\{e\}} \upsilon_{i \leftarrow e'} (c_i)  + \sum_{k} \bar{\alpha}_{i \leftarrow k} ] \nonumber\\
&= \max_{c_i: c_i \neq i}
[
s(i, c_i) + \tilde{\alpha}_{i \leftarrow c_i} (c_i)] + \sum_{e' \in \mathcal{E}^R_i/\{e\}} \upsilon_{i \leftarrow e'} (c_i \neq i)  + \sum_{k} \bar{\alpha}_{i \leftarrow k}.
\end{align}

For $\tilde{\rho}_{i \rightarrow k} (c_i = k)$
and $\tilde{\alpha}_{i \leftarrow k}(c_i = k)$,
we can obtain
\begin{equation}
\tilde{\rho}_{i \rightarrow k} (c_i = k) = \bar{s}(i, k) - \max\nolimits_{i' : i' \neq k} [\bar{s}(i, i') + \alpha_{i \leftarrow k}(c_i = i')].
\end{equation}

\begin{equation}
\bar{s}(i, k) = \left\{
\begin{array}{ll}
\sum_{e \in \mathcal{E}_i} v_{i \leftarrow e}(c_i = i) + s(i, i), & k = i \\
s(i, k), & k \neq i.
\end{array}\right.
\end{equation}

\begin{align}
\tilde{\alpha}_{i \leftarrow k}(c_i = k) = \left\{
\begin{array}{ll}
\sum_{i': i' \neq k} \max(0, \tilde{\rho}_{i'\rightarrow k}(c_i = k)), & k = i \\
\min[0, \tilde{\rho}_{ k \rightarrow k}(c_k = k)) + \sum_{i': i' \neq i, k} \tilde{\rho}_{i'\rightarrow k}(c_i = k))], & k \neq i.
\end{array}\right.
\end{align}

For $\tilde{\upsilon}$ and $\tilde{\pi}$,
we have the following derivations
\begin{align}
&\tilde{\upsilon}_{i\leftarrow e}(c_i = i)  \nonumber\\
=&v_{i\leftarrow e}(c_i = i) - \bar{\upsilon}_{i\leftarrow e} \nonumber\\
=&\max [ p(i, j) + \max\nolimits_{b_j: b_j \neq j} \pi_{j \rightarrow e}(b_j), q(i, j) + \pi_{j \rightarrow e}(j) ]
- \max [ \bar{q}(i,j) + \max\nolimits_{b_j: b_j \neq j} \pi_{j \rightarrow e}(b_j), p(j, i) + \pi_{j \rightarrow e}(j) ] \nonumber\\
=& \max [ p(i, j), q(i, j) + \pi_{j \rightarrow e}(j) - \max\nolimits_{b_j: b_j \neq j} \pi_{j \rightarrow e}(b_j) ]
- \max [ \bar{q}(i,j) + , p(j, i) + \pi_{j \rightarrow e}(j) - \max\nolimits_{b_j: b_j \neq j} \pi_{j \rightarrow e}(b_j) ] \nonumber\\
=& \max [ p(i, j), q(i, j) + \tilde{\pi}_{j \rightarrow e}(j) ]
- \max [ \bar{q}(i,j), p(j, i) + \tilde{\pi}_{j \rightarrow e}(j) ].
\end{align}

\begin{align}
&\tilde{\upsilon}_{i\leftarrow e}(c_i \neq i)
= \upsilon_{i \leftarrow e} (c_i \neq i) - \bar{\upsilon}_{i \leftarrow e} = 0.
\end{align}

\begin{align}
&\tilde{\pi}_{i \rightarrow e} (c_i = i) \nonumber \\
=&\pi_{i \rightarrow e} (c_i = i) - \bar{\pi}_{i \rightarrow e} \nonumber\\
= &s(i, i) + \tilde{\alpha}_{i \leftarrow i} (i)
- \max_{c_i: c_i \neq i} [ s(i, c_i) + \tilde{\alpha}_{i \leftarrow c_i} (c_i) ]
+ \sum_{e' \in \mathcal{E}_i/\{e\}} \tilde{\upsilon}_{i \leftarrow e'} (i).
\end{align}

It can observed that only the variables
$\tilde{\rho}_{i \rightarrow k} (c_i)$ and
$\tilde{\alpha}_{i \leftarrow k} (c_i)$ for $c_i = k$
and
$\tilde{\upsilon}_{i \leftarrow e} (c_i)$
and $\tilde{\pi}_{i \rightarrow e} (c_i)$
for $c_i = i$
are involved in the message passing.
Therefore,
we can define scalar-valued variables
$r(i, k) = \tilde{\rho}_{i \rightarrow k} (c_i = k)$,
$a(i, k) = \tilde{\alpha}_{i \leftarrow k}(c_i = k)$,
$v(i, j) = \tilde{\upsilon}_{i \leftarrow e_{ij}}(c_i = i)$,
and $w(i, j) = \tilde{\pi}_{i \rightarrow e_{ij}}(c_i = i)$.
These scalar messages are summarized as follows.

%\begin{figure} [t]
%\centering
%\subfigure[(a) $\rho_{i\rightarrow k}$]{
%\label{fig:Messages:rho}
%\includegraphics[scale = 1.4]{fig/rho}
%}~~~~~~~~
%\subfigure[(b) $\alpha_{i\leftarrow k}$]{
%\label{fig:Messages:alpha}
%\includegraphics[scale = 1.4]{fig/alpha}
%}~~~~~~~~
%\subfigure[(c) $\pi_{i\rightarrow e}$]{
%\label{fig:Messages:pi}
%\includegraphics[scale = 1.4]{fig/pi}
%}~~~~~~~~
%\subfigure[(d) $\upsilon_{i\leftarrow e}$]{
%\label{fig:Messages:upsilon}
%\includegraphics[scale = 1.4]{fig/upsilon}
%}~~~~~~~~
%\subfigure[(e) belief]{
%\label{fig:Messages:belief}
%\includegraphics[scale = 1.4]{fig/belief}
%}
%\MyCaption{Vector-valued messages.}
%\label{fig:Messages}
%\end{figure}

\mytextbf{Homogeneous message propagation}
There are two kinds of messages
exchanged within image points.
The~\Mytextquotedbl{responsibility}
$r(i,k)$,
sent from data point $i$ to data point $k$,
which reflects how well $k$ serves as the exemplar of $i$
considering other potential exemplars for $i$,
and the~\Mytextquotedbl{availability}
$a(i,k)$,
sent from data point $k$ to data point $i$,
which reflects how appropriately $i$ chooses $k$
as its exemplar considering other potential points that may choose $k$ as their exemplar.
The messages are updated in an iterative way as
\begin{equation}
r(i, k) = {\bar{s}(i, k)} - \max\nolimits_{i': i' \neq k}[{\bar{s}(i, i')} + a(i, i')].
\label{eqn:RMessage}
\end{equation}

\begin{equation}
{
\bar{s}(i, k) = \left\{
\begin{array}{ll}
\sum_{j \in \mathcal{E}^R_{i.}} v(i, j) + s(i, i) & k = i ,\\
s(i, k)& k \neq i .
\end{array}\right.}
\label{eqn:SbarMessage}
\end{equation}

\begin{equation}
a(i, k) = \left\{
\begin{array}{ll}
\sum\nolimits_{i': i' \neq k} \max(0, r(i', k)) & k = i, \\
\min[0, r(k, k) + \sum\nolimits_{i': i' \neq i, k} \max(0, r(i', k))] & k \neq i.
\end{array}\right.
\label{eqn:AMessage}
\end{equation}

\mytextbf{Heterogeneous message propagation}
There are two kinds of message
exchanged between images and words.
The~\Mytextquotedbl{discardability}
$w(i, j)$,
sent from image $i$ to word $j$,
which reflects
how much it is affected that image $i$
selects itself as its exemplar
when the contribution of word $j$ is discarded
and helps word $j$ make better decision
whether to select itself as its exemplar.
The~\Mytextquotedbl{contributability}
$v(i, j)$,
sent from word $j$ to image $i$,
which reflects how well image $i$ serves as an exemplar
considering whether word $j$ is an exemplar.
The two messages are updated as

\begin{align}
w(i, j) = r(i, i) + a(i, i) - v(i, j) = t(i, i) - v(i, j).
\label{eqn:WMessage}
\end{align}

\begin{align}
v(i, j)
=& \max \{ p(i, j), q(i, j) + w(j, i)\} \nonumber \\
&- \max \{\bar{q}(i,j), p(j, i) + w(j, i) \}.
\label{eqn:VMessage}
\end{align}

%\begin{figure} [t]
%\centering
%\subfigure[(a) $r(i, k)$]{
%\label{fig:ScaleMessages:r}
%\includegraphics[scale = 1.4]{fig/r1}
%}~~~~~~~~
%\subfigure[(b) $a(i, k)$]{
%\label{fig:ScaleMessages:a}
%\includegraphics[scale = 1.4]{fig/a1}
%}~~~~~~~~
%\subfigure[(c) $w(i, j)$]{
%\label{fig:ScaleMessages:w}
%\includegraphics[scale = 1.4]{fig/w1}
%}~~~~~~~~
%\subfigure[(d) $v(i, j)$]{
%\label{fig:ScaleMessages:v}
%\includegraphics[scale = 1.4]{fig/v1}
%}
%\MyCaption{Scalar-valued messages.}
%\label{fig:ScaleMessages}
%\end{figure}

\mytextbf{Exemplar assignment}
To obtain exemplar assignment
after convergence,
we sum together
all the incoming messages to $c_i$
and take the value $\hat{c}_i$
as follows:
\begin{align}
\hat{c}_i = &\arg\max\nolimits_{i'} [\sum\nolimits_{i'} \alpha_{i \leftarrow k}(i') + s(i, i') + \sum\nolimits_{j}\upsilon_{i \leftarrow e_{ij}} (i')] \nonumber \\
=& \arg\max\nolimits_{i'} [\sum\nolimits_{k} \tilde{\alpha}_{i \leftarrow k}(i') + \sum\nolimits_{k} \bar{\alpha}_{i \leftarrow k} + s(i, i')
 + \sum\nolimits_{j} \tilde{\upsilon}_{i \leftarrow e_{ij}} (i') + \sum\nolimits_{j} \bar{\upsilon}_{i \leftarrow e_{ij}}] \nonumber \\
=&
\arg\max\nolimits_{i'}
\left\{
\begin{array}{ll}
a(i, i') + s(i, i') & i' \neq i\\
a(i, i') + \sum_{j}v(i, j) + s(i, i') & i' = i
\end{array}\right. \nonumber \\
=&
\arg\max\nolimits_{i'}
[a(i, i') + \bar{s}(i, i')] \nonumber \\
=& \arg\max \nolimits_{i'} [ a(i, i') + \bar{s}(i, i')
-\max_{k: k \neq i'}(a(i, k) + \bar{s}(i, k)] \nonumber \\
=&
\arg\max\nolimits_{i'}
[a(i, i') + r(i, i')].
\end{align}

\subsection*{Complexity analysis}
This section presents the complexity analysis
of the proposed algorithm.
The naive implementation of heterogeneous affinity propagation
would take $O(n^3 + m^3 + mn(m + n))$
per iteration.
In the following,
we analyze the algorithm carefully
and justify
the algorithm essentially
only costs $O( |\mathcal{E}^I| + |\mathcal{E}^W| + |\mathcal{E}^R| )$
per iteration
through
the trick of reusing some computations.
The analysis borrows some ideas from~\cite{FreyD07},
but differs from it
because in our algorithm
the responsibility message involves
the sum of contributability messages
and heterogeneous message propagation
is additionally introduced.

When computing the responsibility message in~\Equation{\ref{eqn:RMessage}},
the maximum and next-to-maximum values
of $\bar{s}(i, i') + a(i, i')$
w.r.t. $i'$
are computed
one time for each $i$
over one pass of the whole algorithm.
Then the maximum value
$\max_{i': i' \neq k}[\bar{s}(i, i') + a(i, i')]$
needed in~\Equation{\ref{eqn:RMessage}},
can be found in a single operation,
by checking to see if $k$ gives the maximum
(in which case the next-to-maximum value is used)
or not
(in which case the maximum value is used).
When computing $\bar{s}(i, k)$,
the summation of $v(i, j)$
w.r.t $j$ is computed
one time for each $i$
over one pass of the whole algorithm.
Similar tricks can also be used
to evaluate $\sum_{i'}\max[0, r(i', k)])$
for computing $a(i, k)$.

All the messages are transmitted
over the edges,
and hence there are totally
$O(|\mathcal{E}^I| + |\mathcal{E}^W| + |\mathcal{E}^R|)$ messages.
The reused computations,
the maximum and next-to-maximum values
of $\bar{s}(i, i') + a(i, i')$,
and the summation of $v(i, j)$
w.r.t $j$,
are performed one time
for each $i$,
which cost $O(|\mathcal{E}^T_i|)$
and $O(|\mathcal{E}^R_{i.}|)$.
Thence the reused computations
cost $O(\sum_i|\mathcal{E}^{I}_i| + \sum_j |\mathcal{E}^{W}_j| +
\sum_i|\mathcal{E}^R_{i.}| + \sum_j|\mathcal{E}^R_{.j}|)
= O(|\mathcal{E}^I| + |\mathcal{E}^W| + |\mathcal{E}^R|)$.
The exemplar assignment for images and words
will cost
$\sum_i|\mathcal{E}^{I}_i| + \sum_j |\mathcal{E}^{W}_j|$.
In summary,
the proposed algorithm costs $O(T'(|\mathcal{E}^I| + |\mathcal{E}^W| + |\mathcal{E}^R|))$
with $T'$ being the iteration number.

 %\flushend
{\footnotesize
\bibliographystyle{ieee}
\bibliography{hap}
}

\end{document}